\documentclass[11pt]{article}
\usepackage[a4paper,margin=1in]{geometry}
\usepackage[T1]{fontenc}
\usepackage{amsmath,amssymb,amsthm}
\usepackage{graphicx,booktabs}
\usepackage[hidelinks]{hyperref}
\usepackage[capitalize,noabbrev]{cleveref}

\newtheorem{theorem}{Theorem}[section]
\newtheorem{proposition}[theorem]{Proposition}
\newtheorem{lemma}[theorem]{Lemma}
\newtheorem{corollary}[theorem]{Corollary}
\theoremstyle{definition}

\newtheorem{remark}[theorem]{Remark}

\newcommand{\R}{\mathbb R}
\newcommand{\T}{\mathsf T}
\DeclareMathOperator{\inertia}{In}

\DeclareMathOperator{\softplus}{softplus}
\DeclareMathOperator{\Var}{Var}

\title{Asymptotic Margins and Feature Confinement in Conic Classification Heads under
Softmax Cross-Entropy}
\author{\"O. N. Gerek\thanks{Department of Electrical and Electronics Engineering,
Eski\c{s}ehir Technical University, Eski\c{s}ehir, Turkey.}}
\date{}

\begin{document}
\maketitle

\begin{abstract}
Conic classification heads can assign bounded acceptance regions to individual classes.
Does softmax cross-entropy then keep features inside a bounded set? For a fixed head whose
scores combine a linear form with a smoothed Euclidean distance, we characterize the loss
of each class as a function of its feature. The asymptotic margin compares the directional
growth of the class score with that of its competitors. The loss is coercive exactly when
the margin is negative, and has infimum zero exactly when it is positive. At zero margin
the loss has a positive floor and remains bounded along at least one ray. When the feature
dimension is at least the number of classes, a second-order cone program gives the
margin. The sign criterion extends to other norms, including those used in polyhedral
conic classifiers. On CIFAR-10 and CIFAR-100, all 1210 class margins of 67 trained conic
heads are positive, including all ten margins in a run constrained to have coercive scores.
For trained weighted prototype heads on CIFAR-10, nine of the ten class losses are coercive.
\end{abstract}


\section{Introduction}
\label{sec:intro}

A conic classifier can reject an unseen sample by thresholding its class scores.
A bounded acceptance region is attractive for this purpose, but what keeps the features
of known classes within it? Consider scores of the form
$\langle w_k,z\rangle+\alpha\|z\|$, with $\alpha>\max_k\|w_k\|$.
Every score grows to infinity with $\|z\|$. Yet the common norm term cancels from the
softmax probabilities, giving exactly the loss of the corresponding linear head.
We ask when this cancellation still permits arbitrarily small loss once each class
has its own centre and norm coefficient.

Conic surfaces have a long history as decision boundaries. The quadratic discriminant of
classical statistics separates two Gaussian classes by a quadric \cite{hastie2009esl}.
The conic section classifier of Banerjee, Kodipaka and Vemuri represents each class by a
focus, a directrix hyperplane and an eccentricity \cite{banerjee2006csc,kodipaka2008margin}.
The polyhedral conic functions of Gasimov and \"Ozt\"urk separate sets that no hyperplane
can separate, using a function built from a linear term and an $\ell_1$ norm about a
centre \cite{gasimov2006pcf}. Bagirov \emph{et al.} turned this idea into a piecewise
linear classifier \cite{bagirov2013pcf}, the incremental conic functions algorithm reduced
its cost on large datasets \cite{cimen2018icf,cimen2018icfsoftx}, and Cevikalp and his
coauthors carried it into object detection, open set recognition and, finally, the
classification stage of a deep network trained end to end
\cite{cevikalp2017pcc,cevikalp2021pami,cevikalp2021deeppcc}.

We study this question for a head built on the norm-linear function of conic
scalarization, which Gasimov and Kasimbeyli introduced for nonconvex vector optimization
\cite{gasimov2001,kasimbeyli2010,kasimbeyli2013}. Its homogeneous zero sublevel set is a cone of the kind
studied by Bishop and Phelps \cite{bishop1963support}, related to the scalarizing
functionals of Gerth and Weidner \cite{gerth1990nonconvex} in the framework of Jahn
\cite{jahn2023unified}. With smoothing parameter $\varepsilon>0$, the norm-linear function of class $k$ reads
\begin{equation}
g_k(z)=\langle w_k,\,z-a_k\rangle+\alpha_k\|z-a_k\|_\varepsilon,
\qquad \|u\|_\varepsilon=\bigl(\|u\|^2+\varepsilon^2\bigr)^{1/2}.
\label{eq:head}
\end{equation}
The classifier assigns to a feature $z$ the class that minimizes $g_k(z)-\rho_k$. Each
score has centre $a_k$, axis $w_k$ and an
aperture governed by $\alpha_k\ge0$. The family contains the affine head as $\alpha_k\to0$
and a prototype head with weighted Euclidean distances as $w_k\to0$. The Euclidean norm
makes the geometry of the head exact, and the main result extends to other norms,
including the $\ell_1$ norm of the polyhedral conic functions.

\subsection{Main results}
\label{sec:results}

For a fixed head, \cref{thm:confine} characterizes the loss of class $k$ by its
asymptotic margin $M_k$, the largest directional advantage of its score over all
competitors. If $M_k>0$, the loss approaches zero only along diverging features and
decreases eventually along each ray with positive directional margin. If $M_k<0$, every
loss sublevel set lies in an explicit ball. At $M_k=0$ the loss has a positive floor but
stays bounded along a ray. The sign criterion also applies to polyhedral conic scores
(\cref{rem:norms}).

Two geometric results explain the distinction. The ratio $\tau_k=\alpha_k/\|w_k\|$
determines the conic type of a single score threshold (\cref{thm:signature}).
The decision cell instead compares all the scores. It contains a ray when $M_k>0$
and is bounded when $M_k<0$. For $d\ge K$, where $d$ is the feature dimension and $K$
the number of classes, \cref{prop:cells} computes $M_k$ by a second-order cone program.

On CIFAR-10 and CIFAR-100 \cite{krizhevsky2009cifar}, all 1210 class margins of the
67 trained conic heads were positive. In the constrained CIFAR-10 run all scores were
coercive, yet every class admitted a direction along which its loss approached zero.
In the prototype runs, nine classes out of ten had negative margins. The experiments
also distinguish confinement from feature scale. The mean test feature norm under the
conic head was 12--38\% smaller than under the affine head on CIFAR-10 and 2\% smaller
on CIFAR-100.

\subsection{Related work and notation}
\label{sec:related}

Three lines of work are related to these results. The first builds geometry into the
head. Heads that normalize both the features and the class weights
\cite{wang2017normface,liu2017sphereface,wang2018cosface,deng2019arcface} divide the
feature space into cones with their apex at the origin. Prototype heads produce Voronoi
cells, and with squared distances they become equivalent to affine heads, because the term
$\|z\|^2$ cancels in the softmax \cite{snell2017proto}. The cells of the norm-linear head
form a curved Voronoi diagram \cite{boissonnat2006curved}, which reduces to the classical
weighted diagrams when all classes share one axis \cite{okabe2000spatial}. The second line
treats the features of the last layer as free variables, in the unconstrained features or
layer-peeled model, in which neural collapse \cite{papyan2020collapse} appears at the
minimizers of a regularized loss with a linear head
\cite{mixon2022neural,fang2021exploring,zhu2021geometric}. \Cref{thm:confine} is a
statement in this model for a fixed conic head and without regularization. It characterizes
when the infimum of the loss is zero and where the small values of the loss lie. The third
line shows that rectifier networks become arbitrarily confident far from the training data
\cite{hein2019relu}, and part (c) of \cref{thm:confine} is the feature-space counterpart
of this far-field confidence. A companion paper studies the same loss as a function of the
parameters of the final layer and of the backbone \cite{gerek2026loss}.

Throughout, $\|\cdot\|$ denotes the Euclidean norm of a vector and the spectral norm of a
matrix, and $[K]=\{1,\dots,K\}$ is the set of classes. A feature is $z\in\R^d$, and a
training set has $n$ samples $x_i$ with labels $y_i$. We write $h_k(z)=g_k(z)-\rho_k$ for
the shifted score of class $k$ and $s_k=-h_k$ for its logit, so softmax cross-entropy of a
feature $z$ with label $y$ is
\begin{equation}
\ell_y(z)=\log\sum_{k=1}^Ke^{s_k(z)}-s_y(z)
=\log\Bigl(1+\sum_{j\ne y}e^{\,h_y(z)-h_j(z)}\Bigr).
\label{eq:ce}
\end{equation}
\Cref{sec:setting} relates the head to existing heads, \cref{sec:head} establishes its
geometry in feature space, and \cref{sec:confine} treats it under the loss.
\Cref{sec:numerics} reports the measurements and \cref{sec:discussion} discusses them.
The supplementary material contains the curvature budget of the head, the proof of
\cref{prop:bisector}, the training protocol, the dynamics of the ratio $\tau_k$ and
further measurements.

\section{Setting and relation to existing heads}
\label{sec:setting}

Existing conic heads use different norms, so we state the condition for boundedness for
an arbitrary norm. Suppose that a class score has the form $\langle w,u\rangle+N(u)$, with
$u=z-a$ and $N$ a norm. Along a ray $u=tv$, this score grows at the rate
$\langle w,v\rangle+N(v)$, so its sublevel sets are bounded exactly when this rate is
positive in every direction, that is, when
\begin{equation}
\|w\|_{N^\ast}=\sup_{N(v)=1}\langle -w,v\rangle<1 ,
\label{eq:dualnorm}
\end{equation}
where $N^\ast$ is the dual norm. In words, the linear term of the score must be weaker
than the norm term in every direction. For the head of \eqref{eq:head}, with
$N=\alpha_k\|\cdot\|$, the condition reads $\tau_k>1$. For the polyhedral conic function,
with $N=\gamma\|\cdot\|_1$, it reads $\|w\|_\infty<\gamma$, and for its extended form, with
$N(v)=\sum_i\gamma_i|v_i|$, it reads $|w_i|<\gamma_i$ for every coordinate. The deep compact
polyhedral conic classifier enforces this last condition through a hinge penalty, and its
authors observe that the acceptance region becomes unbounded when the penalty margin is set
to zero \cite{cevikalp2021deeppcc}. Writing $\tau=1/\|w\|_{N^\ast}$ for a general norm,
$\tau>1$ means bounded sublevel sets for each of these heads. The polyhedral conic
classifiers fix $\tau>1$ by design, because open set recognition requires a bounded
acceptance region, and the conic section classifier learns its eccentricities on fixed
features \cite{banerjee2006csc,kodipaka2008margin}. In the present work, the ratio, the
apices and the apertures are learned together with the backbone, and the value at which
the ratio settles is a measured result of training.

The choice of norm determines which shapes the class regions and the decision boundaries
can take. The $\ell_1$ norm produces acceptance regions shaped as distorted octahedra and
piecewise linear boundaries, whereas the Euclidean norm of \eqref{eq:head} produces the
quadric regions of \cref{thm:signature} and the boundaries of degree at most four of
\cref{prop:bisector}. The focus--directrix definition of a conic uses the Euclidean
distance to the focus, so only the Euclidean norm allows $1/\tau_k$ to be read as an
eccentricity. A Euclidean variant also appears in \cite{cevikalp2021pami}, constrained to
bounded regions.

Since adding the same constant to every logit does not change the softmax probabilities,
only the differences between class scores matter under softmax. Three exact relations
follow. First, if all classes share one apex $a$ and one aperture $\alpha$, then
$g_i-g_j=\langle w_i-w_j,z-a\rangle$ is affine in $z$, and the conic head makes the same
decisions, with the same loss, as an affine head, whatever the ratios $\tau_k$. Curved
decision boundaries therefore arise only when the apices or the apertures differ. Second,
if every axis vanishes, the scores $\alpha_k\|z-a_k\|_\varepsilon$ define a prototype head
with weighted distances. Third, a cosine head depends only on the direction of $z$,
whereas every score of \eqref{eq:head} grows linearly with the distance from its apex, so
neither family contains the other.

\section{Geometry of the norm-linear head}
\label{sec:head}

This section establishes the shape of the region of each class, the decision boundaries
between two classes and the boundedness of the cells. The curvature that the head adds to
the loss is treated in \ref{sm:budget}. In brief, Theorem~\ref{thm:budget} there shows that the
Hessian of each score has norm at most $\alpha_k/\varepsilon$, attained at the apex, and
that the scores grow at least at the rate $\min_k\|w_k\|(\tau_k-1)$, so the smoothing
controls the curvature of the head and the ratio controls the growth of its scores
independently. Combined with softmax cross-entropy, the head enters the Hessian of the
empirical risk only through its Jacobian and its second derivatives, with constants that
do not depend on the number of classes.

\subsection{The conic type of a class region}
\label{sec:type}

We call the set of features at which the score $g_k$ does not exceed the offset $\rho_k$
the region of class $k$. The next theorem, proved in \ref{app:proofs}, shows that the
ratio $\tau_k=\alpha_k/\|w_k\|$ determines its shape.

\begin{theorem}[Conic type of a class region]\label{thm:signature}
Let $u=z-a_k$, $\tilde u=[u^{\T},1]^{\T}$ and $A_k=\alpha_k^2 I-w_kw_k^{\T}$. Then
\begin{equation}\label{eq:quadric}
\{z: g_k(z)\le\rho_k\}
=\{z:\tilde u^{\T}C_k\tilde u\le 0\}\cap\{z:\langle w_k,u\rangle\le\rho_k\},
\end{equation}
where
\begin{equation*}
C_k=\begin{bmatrix}A_k & \rho_k w_k\\ \rho_k w_k^{\T} & \alpha_k^2\varepsilon^2-\rho_k^2\end{bmatrix}.
\end{equation*}
Suppose $d\ge2$, $\varepsilon>0$, $\alpha_k>0$ and $w_k\ne0$. The matrix $A_k$ has the eigenvalue $\alpha_k^2$ with multiplicity $d-1$ and the
eigenvalue $\alpha_k^2-\|w_k\|^2$ once. Its inertia $\inertia(A_k)$, which lists the
numbers of positive, zero and negative eigenvalues, is therefore $(d,0,0)$, $(d-1,1,0)$
or $(d-1,0,1)$ when $\tau_k$ is greater than, equal to or less than unity. The set has a
non-empty interior if and only if $\rho_k>\varepsilon(\alpha_k^2-\|w_k\|^2)^{1/2}$ when
$\tau_k>1$, if and only if $\rho_k>0$ when $\tau_k=1$, and for every $\rho_k$ when
$\tau_k<1$. When the interior is non-empty, the boundary of the set is a quadric of
revolution about the line through $a_k$ parallel to $w_k$, with eccentricity
$e_k=1/\tau_k$. This quadric is a prolate spheroid when $\tau_k>1$, an elliptic
paraboloid when $\tau_k=1$, and one sheet of a two-sheeted hyperboloid when $\tau_k<1$.
For a region with interior, the half-space in \eqref{eq:quadric} removes no point in the first two cases, whereas in
the third case it is always needed. In the nondegenerate unsmoothed case ($\varepsilon=0$, $\rho_k\ne0$), the apex $a_k$ is a focus of the
quadric, and the hyperplane $H_k=\{z:\langle w_k,z-a_k\rangle=\rho_k\}$ is the
corresponding directrix.
\end{theorem}

The theorem has a direct picture in one more dimension. At $\varepsilon=0$, the graph of
the norm term, $y=\alpha_k\|z-a_k\|$, is a right circular cone over the feature space with
its apex at $a_k$. The region of the class is the set of features above which this cone
lies below the plane $y=\rho_k-\langle w_k,z-a_k\rangle$, and its boundary is the
projection of the curve in which the plane cuts the cone. The cone rises with slope
$\alpha_k$ and the plane with slope $\|w_k\|$, so $\tau_k$ compares the two slopes, and the
three cases of the theorem are the three sections of Apollonius. At $\tau_k=1$, the matrix
$A_k$ becomes singular, which here gives the parabolic case, with the region unbounded
along the single direction $-w_k$. A larger $\varepsilon$ shrinks the region and leaves
its axis and eccentricity unchanged, because $\varepsilon$ enters $C_k$ only through the
corner entry and the block $A_k$ does not contain it.

\subsection{Decision boundaries}
\label{sec:bound}

The cell of class $k$ is the set of features that the classifier assigns to it,
\begin{equation}
\mathcal R_k=\{z: h_k(z)\le h_j(z) \text{ for all } j\},
\label{eq:cell}
\end{equation}
and the boundary between two cells lies on the set where the two shifted scores agree.
To describe this set, we write $\nu_k(z)=\|z-a_k\|_\varepsilon$ and
$L_{ij}(z)=\langle w_i,z-a_i\rangle-\langle w_j,z-a_j\rangle-\rho_i+\rho_j$, an affine
function that collects the linear terms and the offsets of the two scores. The set where
the two shifted scores agree is then
$\mathcal B_{ij}=\{z:\alpha_i\nu_i(z)-\alpha_j\nu_j(z)=-L_{ij}(z)\}$.

\begin{proposition}[Taxonomy of the decision boundaries]\label{prop:bisector}
Every point of $\mathcal B_{ij}$ satisfies
\begin{equation}\label{eq:boundary}
\bigl(\alpha_i^2\nu_i^2-\alpha_j^2\nu_j^2-L_{ij}^2\bigr)^2=4\alpha_j^2L_{ij}^2\nu_j^2 ,
\end{equation}
which is a polynomial equation of total degree at most four in $z$. For $\varepsilon,\alpha_i,\alpha_j>0$, when $w_i=w_j$ and $a_i\ne a_j$, the function $L_{ij}\equiv c$ is constant, and the set $\mathcal B_{ij}$,
whenever it is non-empty, is a hyperplane if $\alpha_i=\alpha_j$ and $c=0$; one sheet of
a two-sheeted hyperboloid of revolution about the line through $a_i$ and $a_j$ if
$\alpha_i=\alpha_j$ and $c\ne0$; a hypersphere (possibly a point) if $\alpha_i\ne\alpha_j$ and $c=0$; and a subset of a
quartic hypersurface invariant under rotations about that line if $\alpha_i\ne\alpha_j$ and $c\ne0$.
In the last case, a meridian section of the unsquared bisector satisfies the Cartesian oval equation at $\varepsilon=0$.
\end{proposition}

\begin{figure}[tbhp]\centering
\includegraphics[width=0.44\textwidth]{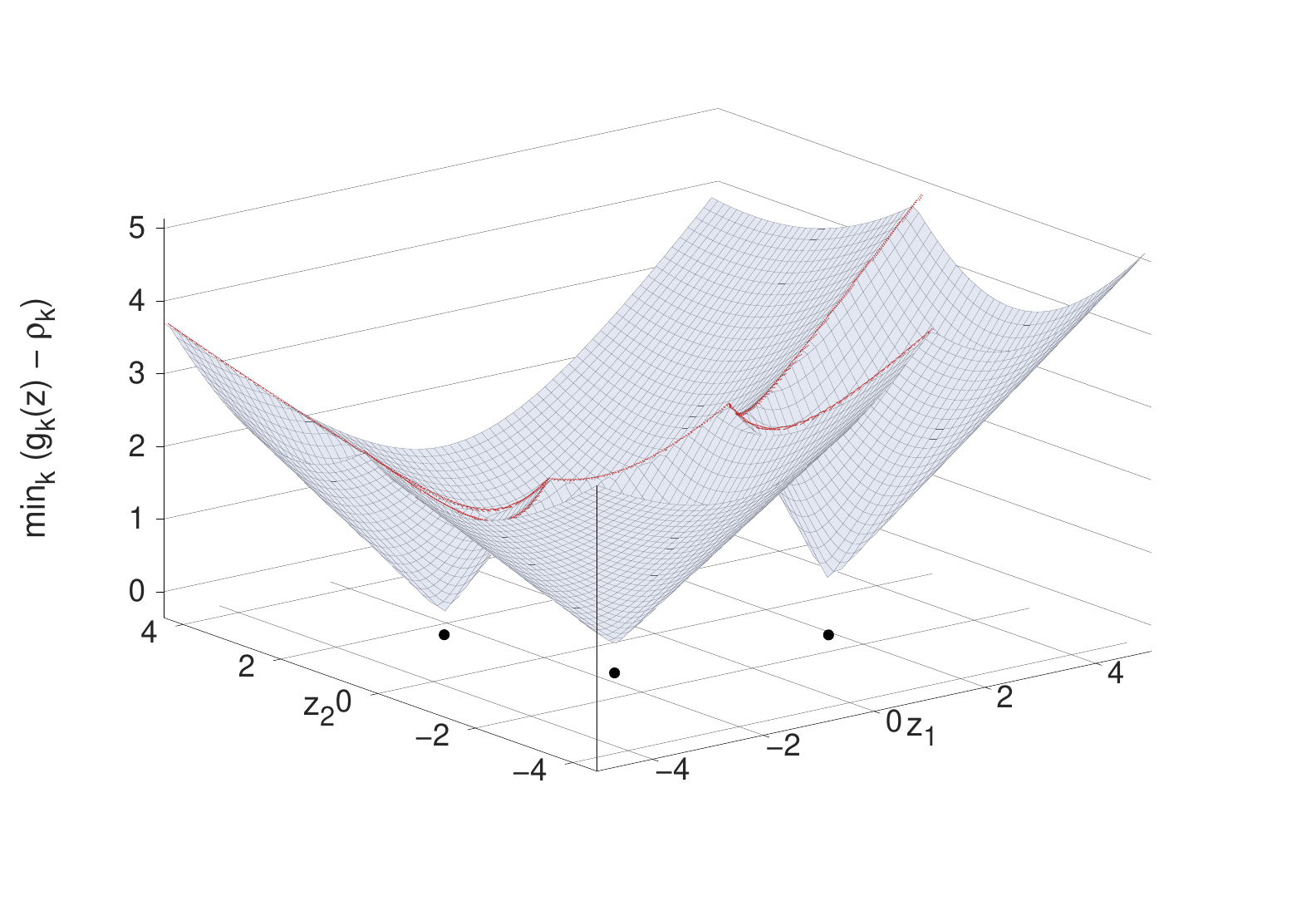}\hfill
\includegraphics[width=0.40\textwidth]{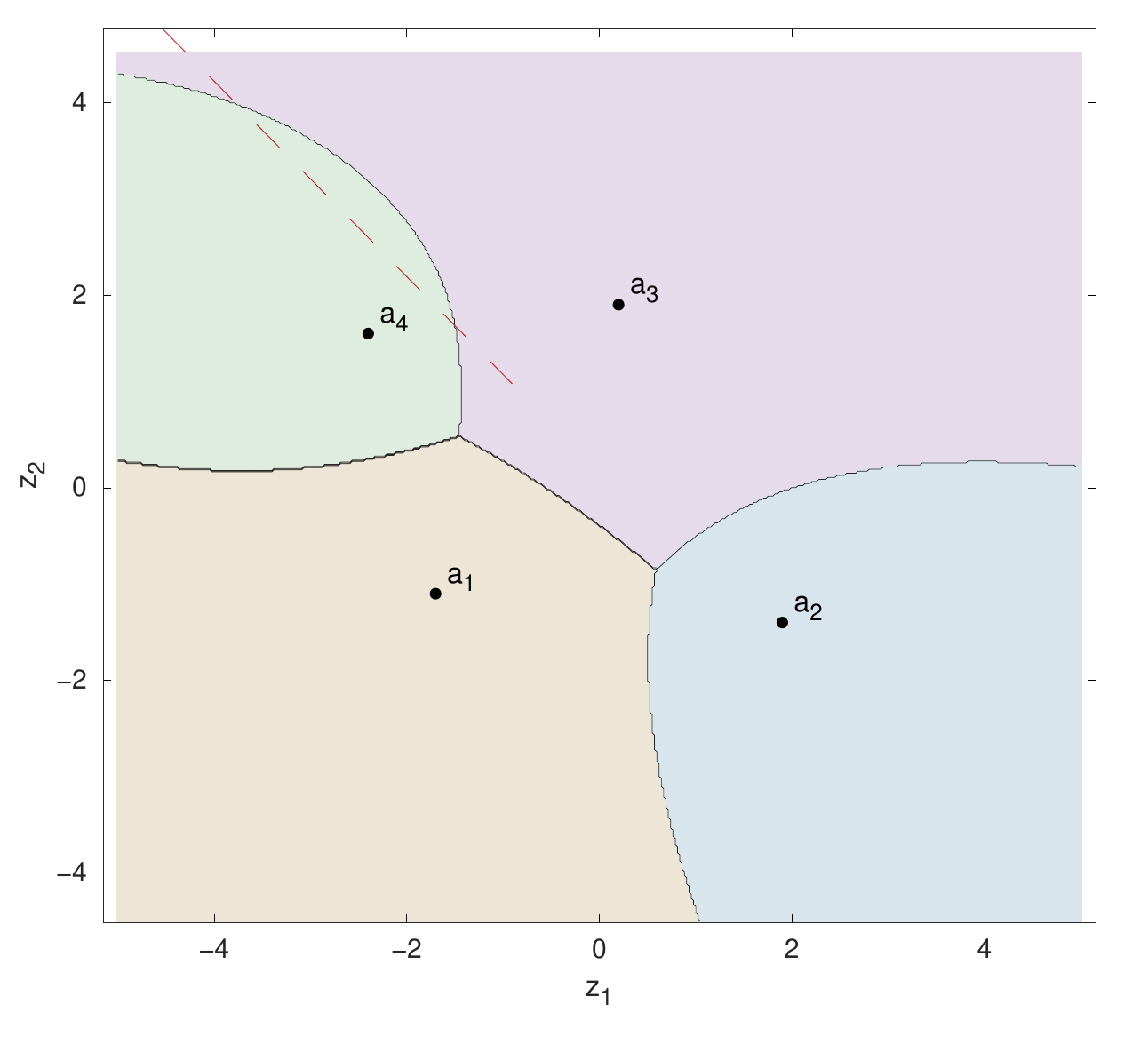}
\caption{The panels illustrate \cref{thm:signature,prop:bisector} with four cones of
unequal apertures and non-parallel axes. The left panel shows the score surface
$\min_kh_k$, the lower envelope of the cones, whose red curves mark the creases along
which the lowest cone changes from one class to another. The right panel shows the same
configuration from above, where the creases appear as curved decision boundaries.}
\label{fig:geometry}
\end{figure}

The proof, by squaring twice and completing squares, is given in \ref{sm:bisector}. The
cells therefore form a curved Voronoi diagram \cite{boissonnat2006curved}. When all
classes share one axis and $\varepsilon=0$, they form the compoundly weighted Voronoi
diagram of the apices, and the four cases of the proposition are the bisectors of its
ordinary, additively, multiplicatively and compoundly weighted forms
\cite{okabe2000spatial}. A flat boundary requires exact equalities between the parameters
of two classes, so flat boundaries do not occur for typical values of the parameters
(\cref{fig:geometry}).

\subsection{Bounded and unbounded cells}
\label{sec:cells}

Along a ray $z=tu$ with $\|u\|=1$, each score grows for large $t$ at the rate
$c_k(u)=\langle w_k,u\rangle+\alpha_k$. We write
\begin{equation}
m_k(u)=\min_{j\ne k}\bigl(c_j(u)-c_k(u)\bigr),\qquad
M_k=\max_{\|u\|=1}m_k(u),
\label{eq:margin}
\end{equation}
and call $M_k$ the asymptotic margin of class $k$. The margin $M_k$ is the largest
amount, over all directions $u$, by which every other score outgrows the score of class
$k$. The terms of the scores that do not grow with the distance are controlled by
\begin{equation}
B_k=\|w_k\|\|a_k\|+\alpha_k(\|a_k\|+\varepsilon),\qquad
E_k=\max_{j\ne k}\bigl(B_j+B_k+|\rho_j-\rho_k|\bigr).
\label{eq:BE}
\end{equation}

\begin{lemma}[Asymptotic expansion of the score differences]
\label{lem:expansion}
For every $z\in\R^d$ and every pair $j\ne k$,
$\bigl|h_j(z)-h_k(z)-\|z\|\,(c_j-c_k)(z/\|z\|)\bigr|\le E_k$, where the middle term is
read as zero at $z=0$.
\end{lemma}

\begin{proof}
Write $g_k(z)=\langle w_k,z\rangle+\alpha_k\|z\|-\langle w_k,a_k\rangle
+\alpha_k\bigl(\|z-a_k\|_\varepsilon-\|z\|\bigr)$. For $z\ne0$, the first two terms equal
$\|z\|\,c_k(z/\|z\|)$. The Cauchy--Schwarz inequality bounds the third term by
$\|w_k\|\|a_k\|$. Since $\|v\|\le\|v\|_\varepsilon\le\|v\|+\varepsilon$, the triangle
inequality gives $\bigl|\|z-a_k\|_\varepsilon-\|z\|\bigr|\le\|a_k\|+\varepsilon$. Therefore
$|g_k(z)-\|z\|c_k(z/\|z\|)|\le B_k$, also at $z=0$, and subtracting the bounds for $g_j$
and $g_k$ and adding the offsets gives the claim.
\end{proof}

\begin{proposition}[Bounded and unbounded cells]\label{prop:cells}
If $M_k<0$, the cell $\mathcal R_k$ is contained in the ball of radius $E_k/|M_k|$ about
the origin. If $M_k>0$ and $u^\ast$ attains the maximum in \eqref{eq:margin}, the cell
contains the ray $\{tu^\ast:t>E_k/M_k\}$. If $S=\operatorname{span}\{w_j-w_k:j\ne k\}$ has dimension below $d$ (in particular if $d\ge K$), then
\begin{equation}
\begin{aligned}
M_k &= \max_{\|u\|\le1}\ \min_{j\ne k}\bigl(\langle w_j-w_k,u\rangle+\alpha_j-\alpha_k\bigr) \\
    &= \min_{\pi\in\Delta}\Bigl(\Bigl\|\sum_{j\ne k}\pi_j(w_j-w_k)\Bigr\|+\sum_{j\ne k}\pi_j(\alpha_j-\alpha_k)\Bigr),
\end{aligned}
\label{eq:dual}
\end{equation}
where $\Delta$ is the probability simplex over the classes $j\ne k$, and the class with
the strictly smallest aperture satisfies $M_k\ge\min_{j\ne k}(\alpha_j-\alpha_k)>0$,
whatever the values of the ratios $\tau_1,\dots,\tau_K$.
\end{proposition}

\begin{proof}
If $M_k<0$ and $\|z\|>E_k/|M_k|$, there is a class $j$ with
$(c_j-c_k)(z/\|z\|)\le M_k$, and \cref{lem:expansion} gives
$h_j(z)-h_k(z)\le\|z\|M_k+E_k<0$, so $z\notin\mathcal R_k$. If $M_k>0$, the same lemma
gives $h_j(tu^\ast)-h_k(tu^\ast)\ge tM_k-E_k>0$ for every $j\ne k$ and $t>E_k/M_k$. For
the dual form, the objective depends on $u$ only through its projection onto $S$, which has dimension at most $K-1$. Any $u$ in the unit ball can
therefore be completed to a unit vector by a component orthogonal to $S$, without changing
the objective, so the maximum over the sphere equals the maximum over the ball. Writing the
inner minimum as a minimum over $\pi\in\Delta$ makes the objective affine in $u$ and in
$\pi$, and the minimax theorem for compact convex sets gives the stated minimum. Finally,
a unit vector $u$ orthogonal to $S$ gives $c_j(u)-c_k(u)=\alpha_j-\alpha_k$ for every $j$.
\end{proof}

Whether a cell is bounded thus depends on how the axes and apertures of its class compare
with those of all the other classes, and the sign of $\tau_k-1$, which concerns one class
alone, cannot by itself decide it. A head in which every ratio exceeds unity still gives
the class of strictly smallest aperture an unbounded cell when $d\ge K$, and a class with
$\tau_k<1$ has a bounded cell as soon as another class has the same axis and a smaller
aperture. The dual form \eqref{eq:dual} is a second-order cone program over $K-1$ weights,
so the sign of $M_k$ can be computed for a trained head. It also gives a simple sufficient
condition for an unbounded cell.

\begin{corollary}[A sufficient condition for an unbounded cell]\label{cor:open}
Let $d\ge K$, and let $\delta_k$ be the distance from $w_k$ to the convex hull of
$\{w_j\}_{j\ne k}$. Then $M_k\ge\delta_k+\min_{j\ne k}(\alpha_j-\alpha_k)$, so the cell
$\mathcal R_k$ is unbounded whenever $\delta_k>\alpha_k-\min_{j\ne k}\alpha_j$.
\end{corollary}

\begin{proof}
For every $\pi\in\Delta$, the vector $\sum_{j\ne k}\pi_jw_j$ lies in the convex hull of
the other axes, so the norm in \eqref{eq:dual} is at least $\delta_k$, and the second sum
is a weighted mean of the differences $\alpha_j-\alpha_k$, so it is at least their minimum.
\end{proof}

When $d\ge K$, the axes are affinely independent for almost all parameter values, so
$\delta_k>0$. For the equiangular configuration of neural collapse
\cite{papyan2020collapse}, in which the axes have a common length $r$ and sum to zero,
$\delta_k=rK/(K-1)$. A cell can therefore be bounded only when the apertures differ by
more than the axes are separated. The prototype head with $w_k=0$ is the extreme case,
with $\delta_k=0$.

\section{Softmax cross-entropy and confinement}
\label{sec:confine}

\Cref{prop:cells} describes the cell when its margin is nonzero. The loss raises a finer question,
namely whether the loss of a class can be made arbitrarily small, and where its small
values lie. The next theorem answers it for the loss \eqref{eq:ce} of a fixed head as a
function of the feature.

\begin{theorem}[Confinement under softmax cross-entropy]
\label{thm:confine}
Let $\ell_y$ be the loss \eqref{eq:ce} of the head \eqref{eq:head} for the label $y$, and
let $m_y$, $M_y$ and $E_y$ be given by \eqref{eq:margin} and \eqref{eq:BE}.
\begin{enumerate}
\item[(a)] If $M_y\le0$, then $\ell_y(z)\ge\log\bigl(1+e^{-E_y}\bigr)$ for every
$z\in\R^d$. If $M_y=0$, the loss also remains bounded along a ray.
\item[(b)] If $M_y<0$, then $\ell_y(z)\ge|M_y|\,\|z\|-E_y$ for every $z$. The loss is
therefore coercive and attains its minimum, and every $z$ with $\ell_y(z)\le\ell_y(z_0)$
lies in the ball of radius $(E_y+\ell_y(z_0))/|M_y|$ about the origin.
\item[(c)] If $M_y>0$, let $u$ be a unit vector with $m_y(u)>0$. Then
\begin{equation}
\ell_y(tu)\le(K-1)\exp\bigl(E_y-t\,m_y(u)\bigr)\qquad\text{for every }t\ge0,
\label{eq:escape}
\end{equation}
and $t\mapsto\ell_y(tu)$ is strictly decreasing on $[t_0(u),\infty)$, where
\begin{equation}
t_0(u)=\max_{j\ne y}\Bigl[\|a_j\|+\Bigl(\frac{\alpha_j(\|a_j\|^2+\varepsilon^2)}{2m_y(u)}\Bigr)^{1/2}\Bigr].
\label{eq:t0}
\end{equation}
The infimum of $\ell_y$ is zero and is not attained, and every sequence $z_n$ with
$\ell_y(z_n)\to0$ satisfies $\|z_n\|\to\infty$.
\end{enumerate}
In particular, $\ell_y$ is coercive if and only if $M_y<0$, and $\inf_z\ell_y(z)=0$ if and only if $M_y>0$.
\end{theorem}

\begin{proof}
By \cref{lem:expansion}, $h_j(z)-h_y(z)\le\|z\|\,(c_j-c_y)(z/\|z\|)+E_y$ for every $z$ and
$j\ne y$, so $\min_{j\ne y}(h_j-h_y)(z)\le\|z\|\,m_y(z/\|z\|)+E_y\le\|z\|M_y+E_y$.

(a) If $M_y\le0$, this gives $\min_{j\ne y}(h_j-h_y)(z)\le E_y$. Keeping only the largest
term of the sum in \eqref{eq:ce} gives
$\ell_y(z)\ge\log\bigl(1+e^{-\min_{j\ne y}(h_j-h_y)(z)}\bigr)\ge\log(1+e^{-E_y})$.

If $M_y=0$, choose $u_*$ with $m_y(u_*)=0$. Then $(c_j-c_y)(u_*)\ge0$ for all
$j\ne y$, so \cref{lem:expansion} gives $h_y(tu_*)-h_j(tu_*)\le E_y$ for $t\ge0$.
Thus $\ell_y(tu_*)\le\log(1+(K-1)e^{E_y})$ along the whole ray.

(b) If $M_y<0$, the same bound gives $\max_{j\ne y}(h_y-h_j)(z)\ge|M_y|\|z\|-E_y$, and
$\log(1+e^x)\ge x$ gives $\ell_y(z)\ge|M_y|\|z\|-E_y$. A continuous function with this
lower bound is coercive and attains its minimum, and $\ell_y(z)\le\ell_y(z_0)$ implies
$|M_y|\|z\|-E_y\le\ell_y(z_0)$.

(c) Let $m=m_y(u)>0$. By \cref{lem:expansion}, $h_y(tu)-h_j(tu)\le-tm+E_y$ for every
$j\ne y$ and $t\ge0$, and $\log(1+x)\le x$ gives \eqref{eq:escape}. The derivative of
$t\mapsto\ell_y(tu)$ is $\sum_{j\ne y}p_j(t)\,\frac{d}{dt}(h_y-h_j)(tu)$, where the weights
$p_j(t)>0$ are the softmax probabilities of the classes $j\ne y$ at $tu$, so it suffices that every derivative in this sum is negative. With
$\varphi_k(t)=\langle tu-a_k,u\rangle/\|tu-a_k\|_\varepsilon\in(-1,1)$, we have
$\frac{d}{dt}h_k(tu)=c_k(u)-\alpha_k(1-\varphi_k(t))$, and therefore
\begin{equation*}
\frac{d}{dt}(h_y-h_j)(tu)=(c_y-c_j)(u)-\alpha_y(1-\varphi_y)+\alpha_j(1-\varphi_j)
\le-m+\alpha_j\bigl(1-\varphi_j(t)\bigr).
\end{equation*}
It remains to show that $\alpha_j(1-\varphi_j(t))<m$ for $t\ge t_0(u)$, which is trivial
when $\alpha_j=0$. Otherwise $t>\|a_j\|$, and with
$\sigma=t-\langle a_j,u\rangle\ge t-\|a_j\|>0$ and $q=\|a_j\|^2-\langle a_j,u\rangle^2
+\varepsilon^2\in(0,\|a_j\|^2+\varepsilon^2]$ we have $\|tu-a_j\|_\varepsilon^2=\sigma^2+q$,
so
\begin{equation*}
1-\varphi_j(t)=\frac{q}{\sqrt{\sigma^2+q}\,\bigl(\sqrt{\sigma^2+q}+\sigma\bigr)}<\frac{q}{2\sigma^2}
\le\frac{\|a_j\|^2+\varepsilon^2}{2(t-\|a_j\|)^2}\le\frac{m}{\alpha_j}.
\end{equation*}
Finally, \eqref{eq:escape} gives $\inf\ell_y=0$, and $\ell_y>0$ everywhere, so the infimum
is not attained. If a sequence with $\ell_y(z_n)\to0$ had a bounded subsequence, a limit
point $\bar z$ would satisfy $\ell_y(\bar z)=0$ by continuity.
\end{proof}

Along each ray with $m_y(u)>0$, the softmax probability $p_y(tu)=e^{-\ell_y(tu)}$
therefore tends to one. This is the feature-space counterpart of the far-field confidence
of rectifier networks \cite{hein2019relu}.

\begin{corollary}[Three heads]\label{cor:heads}
The following hold for the loss \eqref{eq:ce}.
\begin{enumerate}
\item[(a)] For an affine head ($\alpha_k=0$ for every $k$) with $d\ge K$, $M_y=\delta_y$,
the distance from $w_y$ to the convex hull of the other axes. No class is confined in the
sense of part (b) of \cref{thm:confine}, and the loss of every class whose axis lies
outside this hull has infimum zero.
\item[(b)] For a prototype head ($w_k=0$ for every $k$) in any dimension,
$m_y(u)=M_y=\min_{j\ne y}(\alpha_j-\alpha_y)$ for every $u$. The loss of class $y$ is
coercive with a positive minimum when another class has a smaller aperture, it has a
positive floor when its aperture is the smallest and shared with another class, and it
tends to zero along every ray when its aperture is strictly the smallest.
\item[(c)] If $d\ge K$ and class $y$ has the strictly smallest aperture, then $M_y>0$
even when every score is coercive.

\end{enumerate}
\end{corollary}

\begin{proof}
For (a), the dual form \eqref{eq:dual} with $\alpha\equiv0$ is the distance from $w_y$ to
the convex hull of the other axes. For (b), $c_k(u)=\alpha_k$ for every $u$. Part (c) is
the last statement of \cref{prop:cells}. Each part then follows from \cref{thm:confine}.
\end{proof}

\begin{remark}[Other norms]\label{rem:norms}
Parts (a), (b) and the first and last statements of part (c) use only the triangle
inequality and positive homogeneity. They therefore hold for scores
$\langle w_k,z-a_k\rangle+N_k(z-a_k)$, with $N_k$ any norm or its smoothed version
$(N_k^2+\varepsilon^2)^{1/2}$, once $c_k(u)=\langle w_k,u\rangle+N_k(u)$ and
$B_k=\|w_k\|\|a_k\|+N_k(a_k)+\varepsilon$. This covers the polyhedral conic functions.
When the norms are multiples $\alpha_kN$ of one norm and $d\ge K$, the class with the
strictly smallest multiple again has a positive margin. A bounded acceptance region for
every class therefore does not confine the features of a polyhedral conic head trained
with softmax cross-entropy, and a bounded region acts as a confinement only when a
decision uses the score of a class itself, as the open set rule of
\cite{cevikalp2021deeppcc} does.
\end{remark}

\begin{remark}[Unconstrained features]\label{rem:ufm}
In the unconstrained features model \cite{mixon2022neural,fang2021exploring,zhu2021geometric}
the features are free variables. For a fixed conic head, \cref{thm:confine} shows that the
empirical risk $\frac1n\sum_i\ell_{y_i}(z_i)$ over free features has infimum zero if and
only if $M_k>0$ for every class that occurs. When some $M_k\le0$, the risk is at least
$\frac{n_k}{n}\log(1+e^{-E_k})$, where $n_k$ is the number of samples of class $k$. Since
the cells cover $\R^d$, at least one margin is nonnegative.
\end{remark}

\section{Numerical illustrations}
\label{sec:numerics}

This section illustrates the statements of \cref{sec:head,sec:confine} on trained networks.
The trained heads use ResNet-18 \cite{he2016resnet} in
its CIFAR form, trained on CIFAR-10 for $100$ epochs by stochastic gradient descent with
momentum $0.9$, a cosine schedule, batches of $128$ and weight decay $5\times10^{-4}$ on
every parameter. The learning rate is $0.1$ with batch normalization \cite{ioffe2015bn}.
Without it we report the rate $0.01$ of our first protocol and the better rate $0.08$. The
conic head uses $\varepsilon=0.5$ and $\alpha_k=\tau_k\|w_k\|$ with
$\tau_k=\softplus(\eta_k)$, starting from $\tau_k=1.5$. Each setting is repeated over five
seeds, and means are compared by the exact two-sided permutation test over the $252$
divisions of ten runs into two groups of five, whose smallest attainable $p$-value is
$0.0079$. We call these the main runs. A set of additional runs, made on a second machine,
repeats these settings and adds runs with modified heads and on CIFAR-100
(\ref{sm:protocol}).

\subsection{Cells and confinement}
\label{sec:num-cells}

For every class of every trained conic head, the asymptotic margin $M_k$, computed from
its dual form \eqref{eq:dual} with bounds that bracket its exact value, is positive. This
covers all $1210$ cells of the $67$ conic heads of the main and the additional runs, among them the
CIFAR-100 heads and a run in which $\tau_k\ge1$ is enforced. \Cref{cor:open} accounts for
the sign in the heads of the main runs and in the CIFAR-100 heads. In the
CIFAR-10 heads of the main runs, $\delta_k$ lies between $1.03$ and
$1.12$ times $\|w_k\|$, close to the equiangular value $K/(K-1)=1.11$, and in the
CIFAR-100 heads between $0.842$ and $0.944$ times $\|w_k\|$, whereas the aperture of a
class exceeds the smallest other aperture by at most $0.18\|w_k\|$, so the bound of the
corollary is positive for each of these classes.

We measured the directional margin of each test feature at the final checkpoint.
In every main run, at least $99.3\%$ of the
correctly classified test features lie on rays with positive directional margin for their own class, so that $tz$ stays
in the cell of $z$ for all large $t$, and the loss pulls a feature inward, in the sense that
$\frac{d}{dt}\ell_y(tz)>0$ at $t=1$, for at most $0.6\%$ of them. Rescaling every feature by
any factor between $0.5$ and $32$ changes the test accuracy by less than $0.3$ points. The
affine head behaves in the same way, with at least $99.8\%$ of the correctly classified test
features in the asymptotic cone of their own cell, in the main runs. In the exploratory run with $\tau_k=1+\softplus(\eta_k)$, the mean ratio
reached 1.88 and feature norms contracted during training. Nevertheless, all ten final
margins were positive.

In each trained prototype head the smallest aperture was unique. Part (b) of
\cref{cor:heads} therefore predicts coercivity for the other nine classes. Nine of the ten cells are bounded,
and the open cell belongs to the class of smallest aperture. Training brings the apertures
close to each other, so the margins are small, between $-0.31$ and $0.08$, while their
signs remain those of the corollary. About $10\%$ of the correctly classified test
features had positive directional margin, all from the class with the open cell.

\subsection{Feature norm, accuracy and the ratio \texorpdfstring{$\tau$}{tau}}
\label{sec:num-tau}

The trained conic heads have positive margins in every class, yet the features differ in scale across heads. In each of the
three settings in which both heads were trained, the mean norm of the test features is
smaller under the conic head, by $12\%$, $38\%$ and $25\%$ (\cref{tab:acc}), and every
conic run lies below every affine run ($p=0.0079$ in each). The gap between the two largest
logits is about the same under both heads, and with batch normalization a larger gain of
the conic head offsets most of the reduction (\ref{sm:heads}). Mean accuracies differed by at most 0.13 percentage points at the tuned CIFAR-10 rates, and the conic head is ahead only at the
under-tuned rate $0.01$, where it trains faster. On CIFAR-100 the conic head settles at
$\tau=0.611\pm0.002$, reduces the feature norm by $2\%$ only ($p=0.032$) and is about one
point less accurate ($75.85\pm0.58\%$ against $76.97\pm0.49\%$, $p=0.024$). The reduction
grows with $\tau_k$, the weight of the norm term relative to the linear term. When $\eta_k$
is excluded from weight decay, the Spearman correlation between $\tau$ and the feature norm
is $-0.98$, $-0.76$ and $-0.99$ in the three settings, and a prototype head and a cosine
head reduce the norm further. The reduction is therefore not specific to conic geometry,
and these results suggest that it appears whenever a head weakens the reward that the loss
gives for growing features.

\begin{table}[tbhp]\centering
\caption{Final test accuracy (\%) and mean norm of the test features, mean $\pm$ sd over
five seeds, with permutation $p$-values for the accuracy.}\label{tab:acc}
\small
\begin{tabular}{@{}lccccc@{}}
\toprule
norm setting & acc., affine & acc., conic & $p$ & $\|z\|$, affine & $\|z\|$, conic\\
\midrule
batch, lr $0.1$     & $95.03\pm0.14$ & $94.90\pm0.17$ & $0.25$  & $6.01\pm0.01$ & $5.27\pm0.24$\\
no batch, lr $0.01$ & $90.48\pm0.17$ & $91.20\pm0.28$ & $0.008$ & $39.9\pm1.7$  & $24.7\pm0.7$\\
no batch, lr $0.08$ & $93.31\pm0.19$ & $93.26\pm0.20$ & $0.68$  & $26.5\pm0.4$  & $19.9\pm0.6$\\
\bottomrule
\end{tabular}
\end{table}

\Cref{thm:signature} identifies $\tau_k$ as the reciprocal eccentricity of each class
region, and the protocol does not constrain it. Weight decay alone drives $\eta_k$ towards
zero and $\tau_k$ towards $\ln2\approx0.693$, at a speed proportional to the learning rate.
We call the trajectory of $\eta_k$ under weight decay alone, with the logged learning rates
and without the loss, its decay path. \ref{sm:tau} shows that, for the optimizer
used here, $\tau_k$ ends above its decay path exactly when the loss has, on balance, pushed
it upwards.

At the matched rate $0.01$, the two backbones send $\tau$ to opposite sides of the decay
path, which ends at $0.953$. With batch normalization, $\tau$ ends at $0.756\pm0.007$,
between $0.19$ and $0.21$ below the path, and without it at $1.203\pm0.011$, between $0.24$
and $0.26$ above, with every run of one setting separated from every run of the other
($p=0.0079$). The loss therefore moves $\tau$ downwards with normalization and
upwards without it. At the larger rates, both settings end above the path, at
$0.914\pm0.072$ with normalization and $0.844\pm0.012$ without it, after an early fall,
and their spreads differ by a variance ratio of $36.7$ ($p=0.004$ on $F(4,4)$). The side of
unity on which $\tau$ settles is therefore set by the learning rate, the weight decay and
the backbone together. When $\eta_k$ is excluded from weight decay, the observed ratios fall on opposite sides of unity, with $\tau$ ending at $0.120\pm0.025$ and $0.894\pm0.018$ with normalization
and at $2.174\pm0.012$ and $2.28\pm0.54$ without it. The final values are not points of
balance between the loss and weight decay, since $\tau_k$ stops moving when the learning
rate vanishes at the end of the cosine schedule (\ref{sm:tau}).

\section{Discussion and conclusion}
\label{sec:discussion}

The region of a class and the confinement of its features are decided by different
quantities. The ratio $\tau_k$ of one class fixes the conic type of its region, by
\cref{thm:signature}. The asymptotic margin $M_k$, which compares the axes and apertures of
all the classes, decides whether the loss of the class can approach zero and where its
small values lie, by \cref{thm:confine}. All 1210 measured margins were positive, including those of the head trained with
coercive scores. In the prototype heads, the unique smallest aperture gave one class
positive margin and all nine others negative margins.

An open set rule that thresholds a class score directly \cite{cevikalp2021deeppcc}
can therefore reject a feature even when softmax assigns its class probability close
to one. Bounded acceptance regions alone give no guarantee that known-class features
stay inside them.

The head also changes the trained network in one reproducible way, which is the smaller
feature norm, and its mechanism is left for further study. The radial component of the loss
gradient on the training features is smaller under the conic head by $24\%$ with
normalization and by $30\%$ without it at the rate $0.01$, whereas at the rate $0.08$ the
two heads exert the same force ($+1.1\%$, $p=0.86$) while the features remain $25\%$
smaller. A weaker outward force of the loss thus accompanies the reduction in two of the
three settings and leaves the third unexplained. We had also expected the bounded curvature
of Theorem~\ref{thm:budget} to widen the range of stable learning rates. With
normalization, the conic head trains stably up to a rate between $0.5$ and $0.6$, well
above the rate $0.1$ in common use, while the affine head still trains at $2.4$
(\ref{sm:sweep}). We did not track the largest curvature during training, which would be needed to test an edge-of-stability explanation
\cite{cohen2021eos}.

\Cref{thm:signature,thm:confine} are exact, and the measurements that illustrate them use
ResNet-18 on CIFAR-10 and CIFAR-100. How the conic head weakens the reward for growing
features, although the growth shared by all scores cancels under softmax, can be studied
with the same head on further datasets and architectures.

\section*{Declaration of competing interest}
The author declares that he has no known competing financial interests or personal
relationships that could have appeared to influence the work reported in this paper.

\section*{Data availability}
The code that trains the networks and reproduces every number and figure in this paper is
publicly available at \url{https://github.com/ongerek/Conic}.

\section*{Use of generative AI in the writing process}
During the preparation of this work the author used Anthropic's Claude models in order to
develop the code, to check and typeset the mathematical formulations, and to edit the
language. OpenAI's ChatGPT assisted with the mathematical and editorial revision. After using these tools, the author reviewed and edited the content as needed
and takes full responsibility for the content of the publication.

\appendix
\section{Proof of Theorem~\ref{thm:signature}}
\label{app:proofs}

\begin{proof}[Proof of \cref{thm:signature}]
The inequality $g_k(z)\le\rho_k$ can be written as
$\alpha_k\|u\|_\varepsilon\le\rho_k-\langle w_k,u\rangle$. Its left-hand side is strictly
positive, so the inequality implies $\langle w_k,u\rangle\le\rho_k$, and under this
condition it is equivalent to its square,
$\alpha_k^2(\|u\|^2+\varepsilon^2)\le\rho_k^2-2\rho_k\langle w_k,u\rangle+\langle w_k,u\rangle^2$,
which rearranges into \eqref{eq:quadric}. The spectrum of $A_k$ follows because
$w_kw_k^{\T}$ has rank one with nonzero eigenvalue $\|w_k\|^2$.

Since $g_k$ is strictly convex, the set has a non-empty interior exactly when
$\rho_k$ exceeds the infimum of $g_k$. When $\tau_k>1$, the function is coercive, and its
gradient $w_k+\alpha_ku/\|u\|_\varepsilon$ vanishes only at $u=-t_*\,w_k/\|w_k\|$ with
$t_*=\varepsilon\|w_k\|(\alpha_k^2-\|w_k\|^2)^{-1/2}$, where the value is
$\varepsilon(\alpha_k^2-\|w_k\|^2)^{1/2}$. When $\tau_k=1$, the bound
$\|u\|_\varepsilon>\|u\|$ and the Cauchy--Schwarz inequality give $g_k>0$, while along
$u=-t\,w_k/\|w_k\|$ the value $\|w_k\|\bigl((t^2+\varepsilon^2)^{1/2}-t\bigr)$ tends to
zero. When $\tau_k<1$, the same ray gives $g_k\to-\infty$.

For $\tau_k\ne 1$, completing the square about $\bar u_k=-\rho_kw_k/(\alpha_k^2-\|w_k\|^2)$
turns the quadric inequality into $(u-\bar u_k)^{\T}A_k(u-\bar u_k)\le\Lambda_k$ with
$\Lambda_k=\rho_k^2\alpha_k^2/(\alpha_k^2-\|w_k\|^2)-\alpha_k^2\varepsilon^2$. In
coordinates $s$ along $w_k$ and $r$ orthogonal to it, measured from $\bar u_k$, the
boundary reads $(\alpha_k^2-\|w_k\|^2)s^2+\alpha_k^2r^2=\Lambda_k$. For $\tau_k>1$ and a
non-empty interior, the semi-axes are $(\Lambda_k/(\alpha_k^2-\|w_k\|^2))^{1/2}$ and
$(\Lambda_k/\alpha_k^2)^{1/2}$, and the eccentricity of the prolate spheroid is
$(1-(\alpha_k^2-\|w_k\|^2)/\alpha_k^2)^{1/2}=1/\tau_k$. For $\tau_k<1$, the coefficient of
$s^2$ is negative and $\Lambda_k<0$, so the set is the interior of a two-sheeted
hyperboloid with eccentricity $(1+(\|w_k\|^2-\alpha_k^2)/\alpha_k^2)^{1/2}=1/\tau_k$. For
$\tau_k=1$, $A_k$ annihilates $w_k$, and in coordinates measured from $a_k$ the boundary
reads $\alpha_k^2r^2+2\rho_k\|w_k\|\,s+\alpha_k^2\varepsilon^2-\rho_k^2=0$, an elliptic
paraboloid since $\rho_k>0$.

For the half-space, let $\tau_k\ge 1$ and $\rho_k>0$, and let $u$ satisfy the quadric
inequality, so that $\alpha_k\|u\|_\varepsilon\le|\rho_k-\langle w_k,u\rangle|$. If
$\langle w_k,u\rangle>\rho_k$ held, we would get
$\alpha_k\|u\|_\varepsilon<\langle w_k,u\rangle\le\|w_k\|\|u\|\le\alpha_k\|u\|$, which is
impossible. When $\tau_k<1$, the points $u=t\,w_k/\|w_k\|$ with large $t$ satisfy the
quadric inequality and violate the half-space. Finally, at $\varepsilon=0$, the defining
inequality divided by $\alpha_k$ reads
$\|z-a_k\|\le\tau_k^{-1}\bigl(\rho_k-\langle w_k,z-a_k\rangle\bigr)/\|w_k\|$, whose
right-hand side is $1/\tau_k$ times the signed distance from $z$ to $H_k$. This is the
focus--directrix definition of a conic with focus $a_k$, directrix $H_k$ and eccentricity
$1/\tau_k$.
\end{proof}


These supplementary materials contain the curvature budget of the head, the proof of
Proposition~\ref{prop:bisector}, numerical examples of the statements of the paper, the training
protocol, and the measurements that the paper summarizes.

\section{The curvature budget of a head}
\label{sm:budget}

This section examines how a head affects the curvature of the empirical risk, that is,
how sharply the loss bends when the parameters of the network change. The backbone, with
parameters $\theta$, produces the feature $z=\phi_\theta(x)$. A head $s:\R^d\to\R^K$ has
the Jacobian $G(z)=\partial s/\partial z$, and the empirical risk is
$\mathcal L=\frac1n\sum_i\ell\bigl(s(z_i),y_i\bigr)$. The budget of a head consists of
three quantities. Its Lipschitz constant $L_g=\sup_z\|G(z)\|$ is the largest factor by
which the head can stretch a change of the feature. Its curvature
$\beta_g=\sup_z\max_k\|\nabla_z^2s_k(z)\|$ is the largest amount by which any logit bends.
Its coercivity constant $\kappa_g=\min_k\liminf_{\|z\|\to\infty}(-s_k(z))/\|z\|$ is the
smallest rate, over all classes and all directions, at which a class score grows far from
the origin. When $\kappa_g>0$, every score is coercive, which means that it increases
without bound in every direction. An affine head $s(z)=Wz+b$ has
$\kappa_g=-\max_k\|W_k\|<0$, since its score $-\langle W_k,z\rangle-b_k$ decreases without
bound along $W_k$.

\begin{lemma}[Separation of head and backbone]
\label{lem:decomp}
Let $J=\partial\phi_\theta/\partial\theta$ be the Jacobian of the backbone and
$H_\ell=\nabla_s^2\ell$ the Hessian of the loss with respect to the logits, and let a
subscript $i$ mark evaluation at the $i$-th training sample. Then
\begin{equation}
\begin{aligned}
\nabla^2_\theta\mathcal L &= \frac1n\sum_i J_i^{\T}
\Bigl[G_i^{\T}H_{\ell,i}G_i+\sum_k\tfrac{\partial\ell}{\partial s_k}\Big|_i
\nabla_z^2s_k(z_i)\Bigr]J_i \\
&\quad +\frac1n\sum_i\sum_j\bigl[G_i^{\T}\nabla_s\ell_i\bigr]_j\nabla^2_\theta\phi^j_\theta(x_i),
\end{aligned}
\label{eq:decomp}
\end{equation}
in which the head appears only through $G$ and through $\nabla_z^2s_k$.
\end{lemma}

\begin{proof}
By the chain rule, the gradient of the loss on one sample is $J^{\T}G^{\T}\nabla_s\ell$.
When this product is differentiated once more with respect to $\theta$, the product rule
gives one term for each of its three factors. The derivative of $\nabla_s\ell$ gives the
Gauss--Newton term $J^{\T}G^{\T}H_\ell GJ$. The derivative of $G$ gives
$J^{\T}\bigl(\sum_k\frac{\partial\ell}{\partial s_k}\nabla_z^2s_k\bigr)J$, in which the
second derivative acts on the head. The derivative of $J$ gives
$\sum_j[G^{\T}\nabla_s\ell]_j\nabla^2_\theta\phi^j_\theta$, in which the second derivative
acts on the backbone. Averaging over the $n$ samples gives (\ref{eq:decomp}).
\end{proof}

Summed over the samples, the first term in (\ref{eq:decomp}) is the generalized
Gauss--Newton matrix of the network \cite{schraudolph2002ggn,martens2020natural}, and the
lemma splits the remaining part of the Hessian according to whether the second derivative
acts on the head or on the backbone. Whatever form the head takes, it therefore affects
the curvature of the risk through two quantities only. The first is its Jacobian $G$,
which measures how strongly the logits respond to a change of the feature. The other is
the set of second derivatives $\nabla_z^2s_k$, which measure how much the head itself
bends. An affine head removes the second quantity entirely, since a linear map does not
bend, and its first quantity is the weight matrix itself, which training is free to
enlarge. For softmax cross-entropy, the two quantities combine into one bound whose
constants depend neither on the head nor on the number of classes.

\begin{lemma}[Softmax cross-entropy]
\label{lem:ce}
For softmax cross-entropy, $\|H_\ell\|\le\frac12$ and $\|\nabla_s\ell\|_1\le2$, and the
bracketed term $H^{\mathrm{feat}}$ of (\ref{eq:decomp}) satisfies
$\|H^{\mathrm{feat}}\|\le\frac12L_g^2+2\beta_g$. The constant $\frac12$ is attained.
\end{lemma}

\begin{proof}
For a unit vector $v$, we have $v^{\T}H_\ell v=\sum_kp_kv_k^2-(\sum_kp_kv_k)^2=\Var_p(v)$,
the variance of a random variable that takes the value $v_k$ with probability $p_k$.
Popoviciu's inequality bounds such a variance by one quarter of the squared range
$(\max_kv_k-\min_kv_k)^2$. For a unit vector, $(v_i-v_j)^2\le2(v_i^2+v_j^2)\le2$, so the
range is at most $\sqrt2$ and $\|H_\ell\|\le\frac12$. The bound is attained, since
$p=(\frac12,\frac12,0,\dots,0)$ and $v=(e_1-e_2)/\sqrt2$ give $\Var_p(v)=\frac12$. The
gradient of the loss is $\nabla_s\ell=p-e_y$, whose $\ell_1$ norm equals $2(1-p_y)\le2$.
Combining the two bounds gives $\|G^{\T}H_\ell G\|\le\frac12\|G\|^2$, and the triangle
inequality gives $\bigl\|\sum_k(\partial\ell/\partial s_k)\nabla_z^2s_k\bigr\|
\le\|\nabla_s\ell\|_1\max_k\|\nabla_z^2s_k\|\le2\beta_g$.
\end{proof}

The constants $2$ and $\frac12$ are the same for every number of classes. The number of
classes can still enter through $L_g$, the norm of the whole $K\times d$ Jacobian. If every
row of $G$ has norm at most $l$, then $\|G\|\le\|G\|_F\le\sqrt K\,l$, and the factor
$\sqrt K$ is reached when all rows are equal. For the head of (\ref{eq:head}), the row
bounds and the other two quantities of the budget have closed forms.

\begin{theorem}[Budget of the norm-linear head]
\label{thm:budget}
Let $g_k$ be given by (\ref{eq:head}), and write $u=z-a_k$ and $\nu=\|u\|_\varepsilon$.
Then, for every $z$,
\begin{equation}
\begin{aligned}
& \|\nabla_zg_k\|\le\|w_k\|+\alpha_k,\qquad
\|\nabla_z^2g_k\|=\frac{\alpha_k}{\nu}\le\frac{\alpha_k}{\varepsilon},\\ 
& g_k(z)\ge(\alpha_k-\|w_k\|)\|z-a_k\| .
\end{aligned}
\label{eq:budget}
\end{equation}
Here the Hessian norm formula assumes $d\ge2$ (in $d=1$ it is $\alpha_k\varepsilon^2/\nu^3$). The curvature bound is attained at $z=a_k$, and the coercivity bound is attained
asymptotically along a ray of minimum growth, so that
$\kappa_g=\min_k(\alpha_k-\|w_k\|)$.
\end{theorem}

\begin{proof}
Since $\nabla\nu=u/\nu$ and $\|u\|\le\nu$, the first inequality follows from the triangle
inequality. Differentiating once more gives $\nabla^2\nu=\nu^{-1}I-\nu^{-3}uu^{\T}$. In the
direction of $u$, this matrix has the eigenvalue $(\nu^2-\|u\|^2)/\nu^3=\varepsilon^2/\nu^3$,
and on the orthogonal complement it has the eigenvalue $1/\nu$ with multiplicity $d-1$. The
latter is the larger of the two, and since $\nu\ge\varepsilon$ with equality at $u=0$, the
second statement follows, including the claim that the bound is attained. The third
statement follows from $\nu\ge\|u\|$ and the Cauchy--Schwarz inequality, which give
$g_k(z)\ge-\|w_k\|\|u\|+\alpha_k\|u\|$. Take $u=-t\,w_k/\|w_k\|$ if $w_k\ne0$, or any ray otherwise. The score equals
$-\|w_k\|t+\alpha_k(t^2+\varepsilon^2)^{1/2}$, and its ratio to $\|z\|$ tends to
$\alpha_k-\|w_k\|$. Taking the minimum over the classes gives $\kappa_g$.
\end{proof}

The bounds in (\ref{eq:budget}) separate the curvature of the head from the growth of its
scores. The curvature bound $\alpha_k/\varepsilon$ does not involve $w_k$, and the
coercivity constant $\alpha_k-\|w_k\|$ does not involve $\varepsilon$. Changing
$\varepsilon$ therefore changes the curvature of the head without changing how fast its
scores grow, and changing $w_k$ changes that growth without changing the curvature.

\begin{remark}[Curvature at the joins]\label{rem:gap}
The parameter $\varepsilon$ rounds the apex of each cone and leaves the joins between
neighbouring cells sharp, so the lower envelope $\min_kh_k$ is not differentiable along
them. Only the softmax smooths the joins, and its curvature enters (\ref{eq:decomp})
through the Gauss--Newton term, which Lemma~\ref{lem:ce} bounds by $\frac12L_g^2$. The
bound $\alpha_k/\varepsilon$ of Theorem~\ref{thm:budget} therefore describes the curvature
inside each cell, while the joins, where the classifier decides between classes, fall
under the Gauss--Newton term.
\end{remark}

\section{Proof of Proposition~\ref{prop:bisector}}
\label{sm:bisector}

\begin{proof}
On $\mathcal B_{ij}$ we have $\alpha_i\nu_i=\alpha_j\nu_j-L_{ij}$. Squaring gives
$\alpha_i^2\nu_i^2-\alpha_j^2\nu_j^2-L_{ij}^2=-2\alpha_jL_{ij}\nu_j$, and squaring once more
yields (\ref{eq:boundary}). Since $\nu_k^2=\|z-a_k\|^2+\varepsilon^2$ is a polynomial of
degree two, both sides of (\ref{eq:boundary}) have degree at most four. If $w_i=w_j$ and
$\alpha_i=\alpha_j=\alpha$, then $\nu_i^2-\nu_j^2=\|z-a_i\|^2-\|z-a_j\|^2$ is affine in $z$,
because the quadratic terms and the smoothing cancel. For $c=0$, the condition
$\nu_i=\nu_j$ is therefore the perpendicular bisector of the two apices. For $c\ne0$, the
bracket in (\ref{eq:boundary}) is affine and the equation is of degree two. In coordinates
$x$ along $a_j-a_i$ and $r$ across it, measured from the midpoint of the apices, it takes
the form
\begin{equation*}
\Bigl(\frac{4f^2\alpha^2}{c^2}-1\Bigr)x^2-r^2=f^2+\varepsilon^2-\frac{c^2}{4\alpha^2},
\qquad f=\tfrac12\|a_i-a_j\| ,
\end{equation*}
where the unsquared equation requires $|c|/\alpha<2f$. In that case, both the coefficient of $x^2$
and the right-hand side are positive, and the quadric is a two-sheeted hyperboloid of
revolution. The unsquared equation keeps the sheet on which $\nu_i-\nu_j$ has the sign of
$-c$. If $\alpha_i\ne\alpha_j$ and $c=0$, the condition $\alpha_i\nu_i=\alpha_j\nu_j$ is
equivalent to $\alpha_i^2\nu_i^2=\alpha_j^2\nu_j^2$, whose quadratic part is
$(\alpha_i^2-\alpha_j^2)\|z\|^2$, so it describes a hypersphere. Completing the square
shows that its centre is $(\alpha_i^2a_i-\alpha_j^2a_j)/(\alpha_i^2-\alpha_j^2)$ and its
squared radius is
$\alpha_i^2\alpha_j^2\|a_i-a_j\|^2/(\alpha_i^2-\alpha_j^2)^2-\varepsilon^2$, so the
hypersphere is empty when $\varepsilon$ exceeds
$\alpha_i\alpha_j\|a_i-a_j\|/|\alpha_i^2-\alpha_j^2|$. With $\alpha_i\ne\alpha_j$ and
$c\ne0$, the bracket in (\ref{eq:boundary}) keeps the term $(\alpha_i^2-\alpha_j^2)\|z\|^2$,
and the equation is of degree four. It depends on $z$ only through the distances to the
two apices, so it is invariant under rotations about the line joining them. At
$\varepsilon=0$, its section by a plane containing that line is $\alpha_ir_i-\alpha_jr_j=-c$,
a Cartesian oval in the bipolar distances $r_i,r_j$.
\end{proof}

When the axes differ, the set $\mathcal B_{ij}$ still lies on the quartic hypersurface
(\ref{eq:boundary}). As $\alpha_k\to0$ for every $k$, each $g_k$ tends to the affine
function $\langle w_k,z-a_k\rangle$, every difference $g_i-g_j$ becomes affine, and every
$\mathcal B_{ij}$ becomes a hyperplane, so the affine head is the limit $\tau_k\to0$ of the
family.

\section{Numerical examples of the theorems}
\label{sm:examples}

Following the proofs of the statements in the paper and in Sections~\ref{sm:budget}
and~\ref{sm:bisector}, we give numerical examples below. They illustrate each result on
small instances in which every quantity can be computed to high accuracy.

\paragraph{Curvature budget} The decomposition (\ref{eq:decomp}) agrees with a
finite-difference Hessian to a relative error of $1.4\times10^{-7}$. In five dimensions,
the largest value of $\|\nabla_z^2g_k\|$ measured over radii from $10^{-3}$ to $10$ is
$25.9985$, which approaches the value $\alpha_k/\varepsilon=26$ that Theorem~\ref{thm:budget}
gives at the apex.

\paragraph{Conic type and boundaries} Region boundaries sampled in the plane at
$\varepsilon=0.5$ satisfy the quadric of (\ref{eq:quadric}) to a residual below
$7\times10^{-12}$. Their eccentricities, measured in the frame of the axis, agree with
$1/\tau_k$ to within $10^{-6}$ at $\tau_k\in\{1.8,1.2,0.8,0.4\}$. At $\tau_k=1$, the
squared distance from the axis is an affine function of the axial coordinate, as a
paraboloid requires, to a relative residual of $10^{-14}$. The minimum of $g_k$ found by
numerical descent agrees with $\varepsilon(\alpha_k^2-\|w_k\|^2)^{1/2}$ to ten digits. For
$\tau_k>1$ with $\rho_k<0$, the quadric set of (\ref{eq:quadric}) is non-empty, while none
of its sampled points lies in the region of the class, as the proof of
Theorem~\ref{thm:signature} predicts. Implicit polynomial fits to decision boundaries sampled at
$\varepsilon=0.5$ leave relative residuals of order $10^{-12}$ at degree one in the flat
case, at degree two in the hyperboloidal and spherical cases, and at degree four in the
quartic case and in the case of unequal axes. In the last two cases, a fit of degree two
leaves residuals of order $10^{-3}$.

\paragraph{Confinement} We draw three heads with $d=6$ and $K=4$, with $\varepsilon=0.5$,
apices and offsets drawn at random: a prototype head with apertures between $1$ and
$1.6$, a conic head with ratios between $0.3$ and $2$, and a conic head in which class
$0$, with $\tau_0=0.8$, shares its axis with class $1$ of smaller aperture. We compute
$M_y$ by dense sampling of the sphere followed by local refinement. Of the twelve classes,
seven have $M_y<0$, among them class $0$ of the third head ($M_0=-0.90$) and three classes
of the prototype head. For each of them, the quantity $|M_y|\|z\|-E_y-\ell_y(z)$ is negative
at $20\,000$ random points with norms between $10^{-2}$ and $10^3$, as part (b) of
Theorem~\ref{thm:confine} requires, and the minimizer found from eight starting points lies well
inside the ball of the theorem (for example at norm $2.84$ inside the radius $34.4$). The
other five classes have $M_y>0$, including the prototype class of smallest aperture with
$M_y=0.027$. Along the maximizing direction, the loss stays below the bound
(\ref{eq:escape}) at $4000$ values of $t$ between $1$ and $400$, and it decreases at every step
beyond $t_0(u)$ of (\ref{eq:t0}), which lies between $5.2$ and $10.8$.

\paragraph{Sign lemma} For the cosine schedules of the paper, with $391$ steps per epoch
for $100$ epochs, the response of $\eta$ to a unit impulse of the loss gradient is
negative at every later step, for impulses placed at the start, in the middle and near
the end of training, at the rates $0.01$, $0.08$ and $0.1$, as Lemma~\ref{lem:decay} states.
When the condition of the lemma is violated, with a step size of $0.1$ and a weight decay
of $0.5$, the response changes sign $285$ times in $4000$ steps.

\section{Training protocol and statistical tests}
\label{sm:protocol}

All trained heads use ResNet-18 \cite{he2016resnet} in its CIFAR form. Training on
CIFAR-10 \cite{krizhevsky2009cifar} runs for $100$ epochs with stochastic gradient
descent, momentum $0.9$, a cosine schedule, batches of $128$ and weight decay
$5\times10^{-4}$, applied to every parameter, including those of the head. The learning
rate is $0.1$ with batch normalization \cite{ioffe2015bn}. Without batch normalization, the
rate $0.01$ was chosen in our first protocol because $0.1$ diverges, and the rate $0.08$
proves better by two to three points for either head. The conic head uses
$\varepsilon=0.5$ and $\alpha_k=\tau_k\|w_k\|$ with $\tau_k=\softplus(\eta_k)$. All of
$a_k$, $w_k$, $\eta_k$ and $\rho_k$ are learned, starting from
$a_k\sim\mathcal N(0,10^{-2}I)$, $w_k\sim\mathcal N(0,d^{-1}I)$, $\tau_k=1.5$ and
$\rho_k=0$. Each setting is repeated over five seeds. A conic head trained with batch
normalization at the rate $0.01$ serves as a control matched in learning rate to the
unnormalized network.

Differences between the means of two settings are assessed by the exact two-sided
permutation test over the $252$ divisions of ten runs into two groups of five.
Differences between the spreads of two settings are assessed by the variance-ratio test
on $F(4,4)$. All accuracies are final test accuracies, so no selection is made on the
test set.

After the main runs, the additional runs comprise $90$ further networks, trained on a second machine, a
workstation with an NVIDIA RTX PRO 6000 Blackwell GPU, under the same protocol.
Thirty-five of them repeat the affine and conic runs of Tables~\ref{tab:acc} and~\ref{tab:tau} with the
same five seeds. In $20$ runs, the parameters $\eta_k$ are excluded from weight decay, in
the four settings of Table~\ref{tab:tau}. In $10$ runs, every axis is held at $w_k=0$, which
gives the prototype limit of the head, with apertures $\alpha_k=\softplus(\eta_k)$ that
start at $1.5$. In $15$ runs, the head is a cosine head whose scale is learned and starts
at $16$. The remaining $10$ runs train the affine and the conic head on CIFAR-100 with
batch normalization, with five seeds each. The repeated runs reproduce the measurements
of the main runs. Their mean accuracies differ from those of the main runs by at
most $0.16$ points, and the mean values of $\tau$ in the four settings of Table~\ref{tab:tau}
are $0.904$, $0.755$, $1.201$ and $0.854$, against $0.914$, $0.756$, $1.203$ and $0.844$.

\section{Further measurements on trained heads}
\label{sm:heads}

\paragraph{Decisions} The norm terms of the conic head contribute little to each
decision. Their median share of the gap between the shifted scores of the assigned class
and of the runner-up class lies between $6\%$ and $18\%$, and it reaches $41\%$ only when
$\tau_k\ge1$ is enforced. We compute the radial derivative of the loss from the log-odds,
so that it remains exact at large margins.

\paragraph{Feature norm, logit gap and gain} The gap between the two largest logits of a
test feature is about the same under both heads, with means of $8.33$ against $8.01$ with
normalization, and of $20.8$ against $20.6$ at the rate $0.01$ and $19.3$ against $19.9$ at
$0.08$ without it. We call the norm of the gradient of this gap with respect to the
feature the gain of the head. The product of the feature norm and the gain measures how
strongly the gap responds when the feature changes by a fixed fraction of its length.
With normalization, the gain of the conic head is $12.5\%$ larger and offsets most of the
reduction of the norm. The product is $17.3$ against $17.7$ ($p=0.13$), and in the
additional runs it is $2.8\%$ smaller, with every run separated. Without normalization, the gain
is larger by $13.5\%$ and $4.7\%$ only, so the product is $30\%$ and $21\%$ smaller under
the conic head, with every run separated in both settings. Within the normalized network,
the reduction of the norm is largest in the exploratory run in which $\tau_k\ge1$ is
enforced ($4.10$ against $6.01$), and in that run the feature norm logged during training
contracts from $6.11$ to $3.72$.

\paragraph{Calibration and outside images} The reduction leaves the calibration of the
network unchanged at the tuned rates. The expected calibration error
\cite{guo2017calibration} is $0.0280$ against $0.0274$ with normalization ($p=0.54$) and
$0.0494$ against $0.0496$ at the rate $0.08$ without it ($p=0.83$). It is lower under the
conic head only at the under-tuned rate $0.01$ ($0.0676$ against $0.0734$, every run
separated). We pass the test images of CIFAR-100 through the networks trained on CIFAR-10
and compare their feature norms with those of the CIFAR-10 test images. With
normalization, the mean norm of the outside features is $83\%$ of the inside norm under
the conic head and $92\%$ under the affine head, and the corresponding values are $67\%$
against $77\%$ at the rate $0.01$ and $62\%$ against $70\%$ at $0.08$ without
normalization. The area under the ROC curve of the feature norm rises from $0.708$ to
$0.804$ with normalization, from $0.761$ to $0.814$ at the rate $0.01$ and from $0.849$ to
$0.874$ at $0.08$. The maximum softmax probability \cite{hendrycks2017baseline} remains
the better score in these networks, and the two heads differ by less than $0.01$ in its
area under the curve in every setting.

\paragraph{Learned decision boundaries} Proposition~\ref{prop:bisector} makes the flatness of a
trained head's decision boundaries measurable. We sample the boundary between two cells
within random two-dimensional planes and compute the ratio of the smaller to the larger
singular value of the sampled points, taken about their mean. This ratio vanishes when
the sampled curve is a straight line. Across ten trained heads at the rate $0.01$, the
values are $0.0178\pm0.0029$ with batch normalization and $0.0273\pm0.0042$ without it,
compared with $10^{-4}$ on a head driven to the affine limit. The learned boundaries are
therefore nearly flat, deviating from a straight line by two to three percent of their
extent, and they remain clearly distinguishable from the affine limit.

\paragraph{Modified heads} \Cref{tab:further} lists the additional runs on CIFAR-10. All $400$ cells of their $40$
conic heads are unbounded. Excluding
$\eta_k$ from weight decay leaves the accuracy unchanged, with differences of at most
$0.17$ points and $p\ge0.08$ in every setting. The feature norm follows $\tau$. With
normalization at the rate $0.1$, the heads that keep $\tau$ near $0.12$ have features as
large as those of the affine head ($6.09$ against $6.04$, $p=0.063$), whereas the heads
with $\tau$ near $0.90$ reduce the norm by $13\%$. Without normalization, the heads with
$\tau$ above $2$ reduce the norm by $71\%$ in both settings, against $37\%$ and $23\%$ for
the heads trained with weight decay on $\eta_k$. The one run at the rate $0.08$ that ends
at $\tau=1.35$ has a feature norm of $15.3$, between those of the two groups. With
normalization at the rate $0.01$, $\tau$ changes only from $0.755$ to $0.894$, and the
feature norm stays the same ($11.5$ against $11.6$, $p=0.21$). The prototype head reduces
the feature norm by $36\%$ with normalization and by $74\%$ without it, and its accuracy is
equal to or slightly above that of the affine head ($95.13$ against $94.92$, $p=0.16$, and
$93.47$ against $93.21$, $p=0.016$). The cosine head removes the feature norm from the
logits, so the loss has no gradient along the direction of the feature and nothing opposes
weight decay on the scale of the features. Their norm falls to $1.08$ with normalization
and to $0.38$ and $0.25$ without it, which is $18\%$ and $1\%$ of the affine values. The
accuracy of the cosine head equals that of the affine head with normalization and at the
rate $0.08$ ($94.99$ against $94.92$, $p=0.56$, and $93.44$ against $93.21$, $p=0.18$),
and it is higher at the under-tuned rate $0.01$ ($91.79$ against $90.51$, every run
separated). The feature norm separates the CIFAR-10 test images from the CIFAR-100 test
images in the order of the norms, with areas under the ROC curve of $0.914$, $0.868$,
$0.806$ and $0.712$ for the cosine, prototype, conic and affine heads with normalization.
For the maximum softmax probability, every head stays within $0.02$ of the affine value,
except the prototype head without normalization, for which the area falls from $0.872$ to
$0.820$.

\begin{table}[tbhp]\centering
\caption{Additional runs: final test accuracy (\%), mean norm of the test
features, and area under the ROC curve of the feature norm for CIFAR-10 against
CIFAR-100 test images, each the mean over five seeds.}\label{tab:further}
\footnotesize\setlength{\tabcolsep}{4pt}
\begin{tabular}{@{}p{2.5cm}ccc@{\quad}ccc@{\quad}ccc@{}}
\toprule
 & \multicolumn{3}{c}{batch norm, lr $0.1$} & \multicolumn{3}{c}{no batch norm, lr $0.01$}
 & \multicolumn{3}{c}{no batch norm, lr $0.08$}\\
\cmidrule(lr){2-4}\cmidrule(lr){5-7}\cmidrule(l){8-10}
head & acc. & $\|z\|$ & AUC & acc. & $\|z\|$ & AUC & acc. & $\|z\|$ & AUC\\
\midrule
affine                        & 94.92 & 6.04 & 0.712 & 90.51 & 39.4 & 0.767 & 93.21 & 26.2 & 0.847\\
conic                         & 94.88 & 5.26 & 0.806 & 91.09 & 24.8 & 0.816 & 93.32 & 20.1 & 0.875\\
conic, $\eta_k$ \\ without decay & 94.90 & 6.09 & 0.705 & 91.22 & 11.3 & 0.832 & 93.33 & 7.57 & 0.888\\
prototype                     & 95.13 & 3.86 & 0.868 & ---   & ---  & ---   & 93.47 & 6.69 & 0.897\\
cosine                        & 94.99 & 1.08 & 0.914 & 91.79 & 0.38 & 0.867 & 93.44 & 0.25 & 0.897\\
\bottomrule
\end{tabular}
\par\smallskip
{\footnotesize A dash marks a setting that was not run. The standard deviation of the
accuracy is at most $0.28$ points.}
\end{table}

\section{Learning-rate sweep and exploratory runs}
\label{sm:sweep}

\Cref{tab:lr} presents a sweep over the learning rate at $20$ epochs. With batch
normalization, the conic head diverges between $0.5$ and $0.6$, while the affine head
still trains, although poorly, at $2.4$. The range of stable rates of the conic head is
therefore narrower by a factor of at least four. Without normalization, both heads fail
between $0.16$ and $0.32$. After $20$ epochs without normalization, the conic head is the
more accurate at every stable rate, which, given the final accuracies of Table~\ref{tab:acc},
reflects its faster early progress in that setting. With normalization, the affine head
is the more accurate at the rates $0.05$, $0.1$ and $0.2$, and the conic head only at
$0.4$.

\begin{table}[tbhp]\centering
\caption{Final test accuracy after $20$ epochs as a function of the learning rate, with
the final $\tau$ of the conic head.}\label{tab:lr}
\begin{tabular}{@{}rccc@{\qquad}rccc@{}}
\toprule
\multicolumn{4}{c}{batch norm} & \multicolumn{4}{c}{no batch norm}\\
\cmidrule(r){1-4}\cmidrule(l){5-8}
lr & affine & conic & $\tau$ & lr & affine & conic & $\tau$\\
\midrule
0.05 & 0.9212 & 0.8979 & 0.489 & 0.005 & 0.7071 & 0.7377 & 1.456\\
0.1  & 0.9205 & 0.8970 & 0.475 & 0.01  & 0.7740 & 0.8055 & 1.417\\
0.2  & 0.9137 & 0.9101 & 0.588 & 0.02  & 0.8429 & 0.8617 & 1.299\\
0.4  & 0.9017 & 0.9061 & 0.609 & 0.04  & 0.8828 & 0.8859 & 1.062\\
0.5  & ---    & 0.8953 & 0.617 & 0.08  & 0.8861 & 0.8881 & 0.680\\
0.6  & ---    & D      & ---   & 0.16  & 0.8776 & 0.8888 & 0.532\\
0.7  & ---    & D      & ---   & 0.32  & 0.1000 & D      & ---\\
0.8  & 0.8870 & D      & ---   & 0.64  & 0.1000 & 0.1000 & ---\\
1.2  & 0.8210 & ---    & ---   &       &        &        &     \\
1.6  & 0.7754 & ---    & ---   &       &        &        &     \\
2.4  & 0.3142 & ---    & ---   &       &        &        &     \\
\bottomrule
\end{tabular}
\par\smallskip
{\footnotesize In the table, D marks a run whose loss became non-finite, a dash marks a
run that was not made, and $0.1000$ is the accuracy of chance.}
\end{table}

\Cref{tab:abl} lists exploratory runs made before the protocol, each a single
observation. When $\tau_k\ge1$ is enforced, through $\tau_k=1+\softplus(\eta_k)$, the
optimizer moves $\tau$ well past the bound, to $1.88$, and the feature norm contracts
during training. Even so, all ten cells remain unbounded, and $99.7\%$ of the correctly
classified test features had positive directional margin. When $\tau$ is held fixed for the first
five epochs and then released, the early fall of $\tau$ is delayed and still occurs. On
CIFAR-100 with batch normalization, where the training loss does not approach zero,
$\tau$ ends at $0.607$ and all $100$ cells are unbounded. A smaller smoothing parameter,
$\varepsilon=0.1$, leaves both the accuracy and the side of unity on which $\tau$ settles
as in the five-seed runs.

\begin{table}[tbhp]\centering
\caption{Exploratory single-seed runs, on CIFAR-10 unless otherwise stated.}\label{tab:abl}
\begin{tabular}{@{}lcc@{}}
\toprule
configuration & final accuracy & final $\tau$\\
\midrule
affine, batch norm                          & 0.9495 & ---\\
conic $\varepsilon=0.1$, $\tau$ free        & 0.9505 & 0.941\\
conic $\varepsilon=0.5$, $\tau\ge1$         & 0.9474 & 1.884\\
conic $\varepsilon=0.5$, $\tau$ frozen 5 ep & 0.9482 & 0.929\\
\midrule
affine, no batch norm, lr $0.01$            & 0.9045 & ---\\
conic $\varepsilon=0.5$, no batch norm, lr $0.01$ & 0.9126 & 1.209\\
\midrule
conic $\varepsilon=0.5$, CIFAR-100, batch norm & 0.7506 & 0.607\\
\bottomrule
\end{tabular}
\end{table}

\section{Where training places the ratio \texorpdfstring{$\tau$}{tau}}
\label{sm:tau}

The ratio $\tau$ settles at a value set by the loss and by weight decay together. The
following lemma makes the dependence precise, and the rest of this section applies it to
the trained networks.

\begin{lemma}[Sign of the deviation from the decay path]\label{lem:decay}
Let $\eta_{t+1}=\eta_t-\xi_tv_{t+1}$ and $v_{t+1}=\beta v_t+\psi_t+\lambda\eta_t$, with
$v_0=0$, momentum $\beta\in(0,1)$, weight decay $\lambda\ge0$, nonincreasing step sizes
$\xi_t>0$ with $\xi_0\lambda\le(1-\sqrt\beta)^2$, and loss gradients $\psi_t$. Let
$\eta^0_t$ follow the same recursion with $\psi\equiv0$ from the same $\eta_0$. Then
$\eta_T-\eta^0_T=-\sum_{s<T}\Gamma_{T,s}\psi_s$ with weights $\Gamma_{T,s}>0$.
\end{lemma}

\begin{proof}
The recursion is linear, so the deviation is $\sum_{s<T}G_{T,s}\psi_s$, where $G_{T,s}$ is
the deviation $e_T$ produced by a unit impulse $\psi_s=1$. For this impulse, $e_t=0$ for
$t\le s$, the buffer deviation is $1$ at $s+1$, and $e_{s+1}=-\xi_s$. For $t\ge s+1$,
eliminating the buffer through $\xi_{t-1}v_t=e_{t-1}-e_t$ gives
$e_{t+1}=(1+\beta\varrho_t-\xi_t\lambda)e_t-\beta\varrho_te_{t-1}$ with
$\varrho_t=\xi_t/\xi_{t-1}\in(0,1]$. Let $x_t=-e_t$, so that $x_s=0$ and $x_{s+1}>0$. At
$t=s+1$ the recursion gives $x_{s+2}\ge(1-\xi_{s+1}\lambda)x_{s+1}\ge(2\sqrt\beta-\beta)
x_{s+1}\ge\sqrt\beta\,x_{s+1}$. If $x_t\ge\sqrt\beta\,x_{t-1}>0$, then
$x_{t-1}\le x_t/\sqrt\beta$ and
\begin{equation*}
x_{t+1}\ge\bigl(1+\beta\varrho_t-\xi_t\lambda-\sqrt\beta\,\varrho_t\bigr)x_t
\ge\bigl(1-\sqrt\beta+\beta-\xi_0\lambda\bigr)x_t\ge\sqrt\beta\,x_t ,
\end{equation*}
using $\varrho_t\le1$ and $\xi_t\le\xi_0$. Hence $\Gamma_{T,s}=-G_{T,s}=x_T>0$ for
$T>s$.
\end{proof}

Since $\tau=\softplus(\eta)$ is increasing, $\tau_k$ ends above its decay path exactly when
the weighted sum of the loss gradients with respect to $\eta_k$ is negative, that is, when
the loss has, on balance, pushed $\tau_k$ upwards. The condition of the lemma holds here,
since $\xi_0\lambda\le5\times10^{-5}$ and $(1-\sqrt{0.9})^2=2.6\times10^{-3}$.

\begin{figure}[tbhp]\centering
\includegraphics[width=0.98\textwidth]{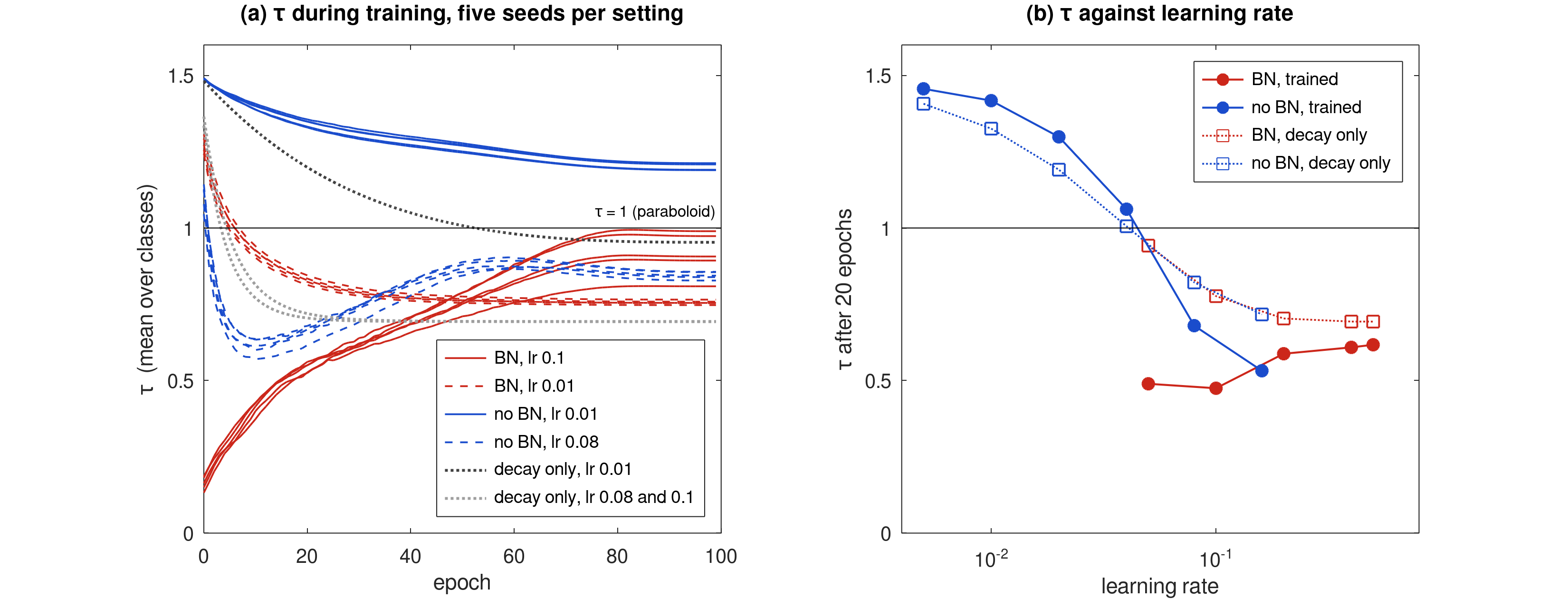}
\caption{The panels show where training places the ratio $\tau$, averaged over the
classes. Panel (a) shows $\tau$ during $100$ epochs in the four settings of
Table~\ref{tab:tau}, five seeds each, together with the decay paths from $\tau=1.5$, drawn
dotted. At the matched rate $0.01$, $\tau$ ends below its decay path with batch
normalization and above it without. At the larger rates, both settings fall early and
then recover above the path. Panel (b) shows $\tau$ after $20$ epochs as a function of the
learning rate (filled markers), together with the decay-only values (open markers).}
\label{fig:tau}
\end{figure}

\Cref{tab:tau} lists the final values of $\tau$ in the four settings of the main protocol,
and \cref{fig:tau} shows their trajectories. The replay of the optimizer on $\eta_k$
alone, with the logged learning rates and without the loss, takes $\tau_k$ from $1.5$ to
$0.693$ at the rates $0.08$ and $0.1$, and to $0.953$ at the rate $0.01$. At the larger
rates, both settings end above the decay path, at $0.914\pm0.072$ with normalization and
$0.844\pm0.012$ without it, after an early fall. With normalization this fall is deep, to
between $0.13$ and $0.19$ within the first epoch, and without normalization it is
moderate, to about $0.6$ near the tenth epoch. The means of the two settings are then no
longer separated ($p=0.056$), whereas their spreads differ by a variance ratio of $36.7$
($p=0.004$). At the matched small rate, the spreads do not differ (ratio $2.96$,
$p=0.32$). The twenty-epoch sweep of \cref{fig:tau}(b) completes the picture. Without
normalization and at rates up to $0.04$, $\tau$ stays within $0.11$ of the value that decay
alone produces, and it crosses unity between $0.04$ and $0.08$. With normalization, $\tau$
lies below the decay path at every rate tried.

When $\eta_k$ is excluded from weight decay, the paths differ as well. With normalization
at the rate $0.1$, $\tau$ falls to $0.13$ within the first epoch and stays near $0.12$
until the end, whereas without normalization at the rate $0.01$ it rises throughout
training. At the rate $0.08$ without normalization, $\tau$ first falls below unity, to
about $0.7$ near the twelfth epoch, as it does under weight decay, and then rises above
unity. Four of these five runs end between $2.27$ and $2.65$, and one ends at $1.35$.

\begin{table}[tbhp]\centering
\caption{Final $\tau$ of the conic head over five seeds per setting, with the value at the
end of the decay path from $\tau=1.5$.}\label{tab:tau}
\begin{tabular}{@{}lccc@{}}
\toprule
setting & $\tau$, mean $\pm$ sd & range & decay path\\
\midrule
batch norm, lr $0.1$     & $0.914\pm0.072$ & $0.809$--$0.990$ & $0.693$\\
batch norm, lr $0.01$    & $0.756\pm0.007$ & $0.747$--$0.765$ & $0.953$\\
no batch norm, lr $0.01$ & $1.203\pm0.011$ & $1.190$--$1.212$ & $0.953$\\
no batch norm, lr $0.08$ & $0.844\pm0.012$ & $0.828$--$0.856$ & $0.693$\\
\bottomrule
\end{tabular}
\end{table}

\paragraph{Weights of the sign lemma} For the schedules of the paper, with $391$ steps per
epoch, \cref{tab:weights} gives the share of the total weight $\sum_s\Gamma_{T,s}$ of
Lemma~\ref{lem:decay} that falls into each quarter of training. At the rate $0.01$, weight
decay changes $\eta$ slowly, and the early and middle epochs carry most of the weight. At
the larger rates, weight decay erases most of the influence of the early gradients, and
the third quarter carries the largest share.

\begin{table}[tbhp]\centering
\caption{Share of the weights $\Gamma_{T,s}$ of Lemma~\ref{lem:decay} in each quarter of
training.}\label{tab:weights}
\begin{tabular}{@{}lcccc@{}}
\toprule
initial rate & epochs 0--24 & 25--49 & 50--74 & 75--99\\
\midrule
$0.01$ & 0.353 & 0.382 & 0.225 & 0.041\\
$0.08$ & 0.015 & 0.217 & 0.582 & 0.186\\
$0.1$  & 0.005 & 0.156 & 0.612 & 0.227\\
\bottomrule
\end{tabular}
\end{table}

\paragraph{No balance at the end of training} At the final checkpoint, we compute the
gradient of the training loss with respect to each $\eta_k$, with all other parameters
held fixed. Every trained $\eta_k$ is positive, so a balance would require the loss to
push each $\eta_k$ upwards with a force equal to the pull $\lambda\eta_k$ of weight decay.
In every five-seed run, however, the loss pushes between two and six of the ten classes
downwards. For the classes that it pushes upwards, its push lies between $0.08$ and $13.7$
times $\lambda\eta_k$, with median $1.5$, and only $12\%$ of these classes lie within a
factor of $1.25$ of a balance. Averaged over the classes, the pushes of the loss largely
cancel, and their mean is between $0.6\%$ and $42\%$ of the mean pull of weight decay. The
same holds when the images are augmented as in training. The ratio $\tau_k$ therefore
stops moving because the learning rate vanishes at the end of the cosine schedule, and its
final value records the path that training took.

\bibliographystyle{abbrv}
\bibliography{references,companion_refs}

@unpublished{gerek2026loss,
  author = {Gerek, {\"O}. N.},
  title  = {Conic Structure of the Cross-Entropy Loss at the Final Layer and Its Absence in Rectifier Backbones},
  year   = {2026},
  note   = {arXiv:\href{https://arxiv.org/abs/2611.20696}{2611.20696} [cs.LG]}
}

@article{bagirov2013pcf,
author    = {Bagirov, Adil M. and Ugon, Julien and Webb, Dean and {\"O}zt{\"u}rk, G{\"u}rkan and Kasimbeyli, Refail},
title     = {A Novel Piecewise Linear Classifier Based on Polyhedral Conic and Max--Min Separabilities},
journal   = {TOP},
volume    = {21},
number    = {1},
pages     = {3--24},
year      = {2013},
publisher = {Springer},
doi       = {10.1007/s11750-011-0241-5}
}

@inproceedings{banerjee2006csc,
author    = {Banerjee, Arunava and Kodipaka, Srujana and Vemuri, Baba C.},
title     = {A Conic Section Classifier and Its Application to Image Datasets},
booktitle = {Proceedings of the IEEE Computer Society Conference on Computer Vision and Pattern Recognition (CVPR)},
volume    = {1},
pages     = {103--108},
year      = {2006},
publisher = {IEEE},
doi       = {10.1109/CVPR.2006.20}
}

@incollection{bishop1963support,
author    = {Bishop, Errett and Phelps, Robert R.},
title     = {The Support Functionals of a Convex Set},
booktitle = {Convexity},
editor    = {Klee, Victor L.},
series    = {Proceedings of Symposia in Pure Mathematics},
volume    = {7},
pages     = {27--35},
publisher = {American Mathematical Society},
address   = {Providence, RI},
year      = {1963},
doi       = {10.1090/pspum/007/0154092}
}

@incollection{boissonnat2006curved,
author    = {Boissonnat, Jean-Daniel and Wormser, Camille and Yvinec, Mariette},
title     = {Curved {Voronoi} Diagrams},
booktitle = {Effective Computational Geometry for Curves and Surfaces},
editor    = {Boissonnat, Jean-Daniel and Teillaud, Monique},
series    = {Mathematics and Visualization},
pages     = {67--116},
publisher = {Springer},
address   = {Berlin, Heidelberg},
year      = {2006},
doi       = {10.1007/978-3-540-33259-6_2}
}

@inproceedings{cevikalp2017pcc,
author    = {Cevikalp, Hakan and Triggs, Bill},
title     = {Polyhedral Conic Classifiers for Visual Object Detection and Classification},
booktitle = {Proceedings of the IEEE Conference on Computer Vision and Pattern Recognition (CVPR)},
pages     = {4114--4122},
year      = {2017},
publisher = {IEEE},
doi       = {10.1109/CVPR.2017.438}
}

@article{cevikalp2021pami,
author    = {Cevikalp, Hakan and Sa{\u{g}}lamlar, Halil},
title     = {Polyhedral Conic Classifiers for Computer Vision Applications and Open Set Recognition},
journal   = {IEEE Transactions on Pattern Analysis and Machine Intelligence},
volume    = {43},
number    = {2},
pages     = {608--622},
year      = {2021},
publisher = {IEEE},
doi       = {10.1109/TPAMI.2019.2934455}
}

@article{cevikalp2021deeppcc,
author    = {Cevikalp, Hakan and Uzun, Burak and K\"{o}p\"{u}kl\"{u}, Okan and \"{O}zt\"{u}rk, G\"{u}rkan},
title     = {Deep Compact Polyhedral Conic Classifier for Open and Closed Set Recognition},
journal   = {Pattern Recognition},
volume    = {119},
pages     = {108080},
year      = {2021},
publisher = {Elsevier},
doi       = {10.1016/j.patcog.2021.108080}
}

@article{cimen2018icf,
author = {\c{C}imen, Emre and \"{O}zt\"{u}rk, G\"{u}rkan and Gerek, \"{O}mer Nezih},
title = {Incremental conic functions algorithm for large scale classification problems},
Journal = {Digital Signal Processing},
volume = {77}, 
Pages = {187--194}, 
Year = {2018},
Doi = {10.1016/j.dsp.2017.11.010}
}

@article{cimen2018icfsoftx,
Author = {\c{C}imen, Emre and \"{O}zt\"{u}rk, G\"{u}rkan and Gerek, \"{O}mer Nezih},
Title = {ICF: An algorithm for large scale classification with conic functions},
Journal =  {SoftwareX}, 
Volume = {8}, 
Pages = {59--63}, 
Year = {2018}, 
Doi = {10.1016/j.softx.2017.12.003}
}

@inproceedings{cohen2021eos,
author    = {Cohen, Jeremy M. and Kaur, Simran and Li, Yuanzhi and Kolter, J. Zico and Talwalkar, Ameet},
title     = {Gradient Descent on Neural Networks Typically Occurs at the Edge of Stability},
booktitle = {International Conference on Learning Representations (ICLR)},
year      = {2021},
}

@inproceedings{deng2019arcface,
author    = {Deng, Jiankang and Guo, Jia and Xue, Niannan and Zafeiriou, Stefanos},
title     = {{ArcFace}: Additive Angular Margin Loss for Deep Face Recognition},
booktitle = {Proceedings of the IEEE/CVF Conference on Computer Vision and Pattern Recognition (CVPR)},
pages     = {4685--4694},
year      = {2019},
publisher = {IEEE},
doi       = {10.1109/CVPR.2019.00482}
}

@incollection{gasimov2001,
author    = {Gasimov, Rafail N.},
title     = {Characterization of the {Benson} Proper Efficiency and Scalarization in Nonconvex Vector Optimization},
booktitle = {Multiple Criteria Decision Making in the New Millennium},
editor    = {K{\"o}ksalan, Murat and Zionts, Stanley},
series    = {Lecture Notes in Economics and Mathematical Systems},
volume    = {507},
pages     = {189--198},
publisher = {Springer},
address   = {Berlin, Heidelberg},
year      = {2001},
doi       = {10.1007/978-3-642-56680-6_17}
}

@article{gasimov2006pcf,
author    = {Gasimov, Rafail N. and \"{O}zt\"{u}rk, G\"{u}rkan},
title     = {Separation via Polyhedral Conic Functions},
journal   = {Optimization Methods and Software},
volume    = {21},
number    = {4},
pages     = {527--540},
year      = {2006},
publisher = {Taylor \& Francis},
doi       = {10.1080/10556780600723252}
}

@article{gerth1990nonconvex,
author    = {Gerth, Christiane and Weidner, Petra},
title     = {Nonconvex Separation Theorems and Some Applications in Vector Optimization},
journal   = {Journal of Optimization Theory and Applications},
volume    = {67},
number    = {2},
pages     = {297--320},
year      = {1990},
publisher = {Springer},
doi       = {10.1007/BF00940478}
}

@inproceedings{guo2017calibration,
author    = {Guo, Chuan and Pleiss, Geoff and Sun, Yu and Weinberger, Kilian Q.},
title     = {On Calibration of Modern Neural Networks},
booktitle = {Proceedings of the 34th International Conference on Machine Learning (ICML)},
editor    = {Precup, Doina and Teh, Yee Whye},
series    = {Proceedings of Machine Learning Research},
volume    = {70},
pages     = {1321--1330},
year      = {2017},
publisher = {PMLR},
}

@book{hastie2009esl,
author    = {Hastie, Trevor and Tibshirani, Robert and Friedman, Jerome},
title     = {The Elements of Statistical Learning: Data Mining, Inference, and Prediction},
edition   = {2nd},
series    = {Springer Series in Statistics},
publisher = {Springer},
address   = {New York, NY},
year      = {2009},
isbn      = {978-0-387-84857-0},
doi       = {10.1007/978-0-387-84858-7}
}

@inproceedings{he2016resnet,
author    = {He, Kaiming and Zhang, Xiangyu and Ren, Shaoqing and Sun, Jian},
title     = {Deep Residual Learning for Image Recognition},
booktitle = {Proceedings of the IEEE Conference on Computer Vision and Pattern Recognition (CVPR)},
pages     = {770--778},
year      = {2016},
publisher = {IEEE},
doi       = {10.1109/CVPR.2016.90}
}

@inproceedings{hendrycks2017baseline,
author    = {Hendrycks, Dan and Gimpel, Kevin},
title     = {A Baseline for Detecting Misclassified and Out-of-Distribution Examples in Neural Networks},
booktitle = {International Conference on Learning Representations (ICLR)},
year      = {2017},
}

@inproceedings{ioffe2015bn,
author    = {Ioffe, Sergey and Szegedy, Christian},
title     = {Batch Normalization: Accelerating Deep Network Training by Reducing Internal Covariate Shift},
booktitle = {Proceedings of the 32nd International Conference on Machine Learning (ICML)},
editor    = {Bach, Francis and Blei, David},
series    = {Proceedings of Machine Learning Research},
volume    = {37},
pages     = {448--456},
year      = {2015},
publisher = {PMLR},
}

@article{jahn2023unified,
author    = {Jahn, Johannes},
title     = {A Unified Approach to {Bishop--Phelps} and Scalarizing Functionals},
journal   = {Journal of Applied and Numerical Optimization},
volume    = {5},
number    = {1},
pages     = {5--25},
year      = {2023},
doi       = {10.23952/jano.5.2023.1.02}
}

@article{kasimbeyli2010,
author    = {Kasimbeyli, Refail},
title     = {A Nonlinear Cone Separation Theorem and Scalarization in Nonconvex Vector Optimization},
journal   = {SIAM Journal on Optimization},
volume    = {20},
number    = {3},
pages     = {1591--1619},
year      = {2010},
publisher = {SIAM},
doi       = {10.1137/070694089}
}

@article{kasimbeyli2013,
author    = {Kasimbeyli, Refail},
title     = {A Conic Scalarization Method in Multi-Objective Optimization},
journal   = {Journal of Global Optimization},
volume    = {56},
number    = {2},
pages     = {279--297},
year      = {2013},
publisher = {Springer},
doi       = {10.1007/s10898-011-9789-8}
}

@inproceedings{kodipaka2008margin,
author    = {Kodipaka, Srujana and Banerjee, Arunava and Vemuri, Baba C.},
title     = {Large Margin Pursuit for a Conic Section Classifier},
booktitle = {Proceedings of the IEEE Conference on Computer Vision and Pattern Recognition (CVPR)},
pages     = {1--6},
year      = {2008},
publisher = {IEEE},
doi       = {10.1109/CVPR.2008.4587406}
}

@techreport{krizhevsky2009cifar,
author      = {Krizhevsky, Alex},
title       = {Learning Multiple Layers of Features from Tiny Images},
institution = {University of Toronto},
address     = {Toronto, ON, Canada},
year        = {2009},
}

@inproceedings{liu2017sphereface,
author    = {Liu, Weiyang and Wen, Yandong and Yu, Zhiding and Li, Ming and Raj, Bhiksha and Song, Le},
title     = {{SphereFace}: Deep Hypersphere Embedding for Face Recognition},
booktitle = {Proceedings of the IEEE Conference on Computer Vision and Pattern Recognition (CVPR)},
pages     = {6738--6746},
year      = {2017},
publisher = {IEEE},
doi       = {10.1109/CVPR.2017.713}
}

@article{martens2020natural,
author    = {Martens, James},
title     = {New Insights and Perspectives on the Natural Gradient Method},
journal   = {Journal of Machine Learning Research},
volume    = {21},
number    = {146},
pages     = {1--76},
year      = {2020},
}

@book{okabe2000spatial,
author    = {Okabe, Atsuyuki and Boots, Barry and Sugihara, Kokichi and Chiu, Sung Nok},
title     = {Spatial Tessellations: Concepts and Applications of {Voronoi} Diagrams},
edition   = {2nd},
series    = {Wiley Series in Probability and Statistics},
publisher = {John Wiley \& Sons},
address   = {Chichester, U.K.},
year      = {2000},
isbn      = {978-0-471-98635-5},
doi       = {10.1002/9780470317013}
}

@article{papyan2020collapse,
author    = {Papyan, Vardan and Han, X. Y. and Donoho, David L.},
title     = {Prevalence of Neural Collapse During the Terminal Phase of Deep Learning Training},
journal   = {Proceedings of the National Academy of Sciences of the United States of America},
volume    = {117},
number    = {40},
pages     = {24652--24663},
year      = {2020},
publisher = {National Academy of Sciences},
doi       = {10.1073/pnas.2015509117}
}

@article{schraudolph2002ggn,
author    = {Schraudolph, Nicol N.},
title     = {Fast Curvature Matrix-Vector Products for Second-Order Gradient Descent},
journal   = {Neural Computation},
volume    = {14},
number    = {7},
pages     = {1723--1738},
year      = {2002},
publisher = {MIT Press},
doi       = {10.1162/08997660260028683}
}

@inproceedings{snell2017proto,
author    = {Snell, Jake and Swersky, Kevin and Zemel, Richard S.},
title     = {Prototypical Networks for Few-Shot Learning},
booktitle = {Advances in Neural Information Processing Systems (NIPS)},
editor    = {Guyon, Isabelle and von Luxburg, Ulrike and Bengio, Samy and Wallach, Hanna and Fergus, Rob and Vishwanathan, S. V. N. and Garnett, Roman},
volume    = {30},
pages     = {4077--4087},
year      = {2017},
publisher = {Curran Associates, Inc.},
}

@inproceedings{wang2017normface,
author    = {Wang, Feng and Xiang, Xiang and Cheng, Jian and Yuille, Alan L.},
title     = {{NormFace}: {$L_2$} Hypersphere Embedding for Face Verification},
booktitle = {Proceedings of the 25th ACM International Conference on Multimedia (ACM MM)},
pages     = {1041--1049},
year      = {2017},
publisher = {ACM},
address   = {New York, NY, USA},
doi       = {10.1145/3123266.3123359}
}

@inproceedings{wang2018cosface,
author    = {Wang, Hao and Wang, Yitong and Zhou, Zheng and Ji, Xing and Gong, Dihong and Zhou, Jingchao and Li, Zhifeng and Liu, Wei},
title     = {{CosFace}: Large Margin Cosine Loss for Deep Face Recognition},
booktitle = {Proceedings of the IEEE/CVF Conference on Computer Vision and Pattern Recognition (CVPR)},
pages     = {5265--5274},
year      = {2018},
publisher = {IEEE},
doi       = {10.1109/CVPR.2018.00552}
}

@article{fang2021exploring,
author    = {Fang, Cong and He, Hangfeng and Long, Qi and Su, Weijie J.},
title     = {Exploring Deep Neural Networks via Layer-Peeled Model: Minority Collapse in Imbalanced Training},
journal   = {Proceedings of the National Academy of Sciences of the United States of America},
volume    = {118},
number    = {43},
pages     = {e2103091118},
year      = {2021},
doi       = {10.1073/pnas.2103091118}
}

@inproceedings{hein2019relu,
author    = {Hein, Matthias and Andriushchenko, Maksym and Bitterwolf, Julian},
title     = {Why {ReLU} Networks Yield High-Confidence Predictions Far Away from the Training Data and How to Mitigate the Problem},
booktitle = {Proceedings of the IEEE/CVF Conference on Computer Vision and Pattern Recognition (CVPR)},
pages     = {41--50},
year      = {2019},
publisher = {IEEE},
doi       = {10.1109/CVPR.2019.00013}
}

@article{mixon2022neural,
author    = {Mixon, Dustin G. and Parshall, Hans and Pi, Jianzong},
title     = {Neural Collapse with Unconstrained Features},
journal   = {Sampling Theory, Signal Processing, and Data Analysis},
volume    = {20},
pages     = {11},
year      = {2022},
doi       = {10.1007/s43670-022-00027-5}
}

@inproceedings{zhu2021geometric,
author    = {Zhu, Zhihui and Ding, Tianyu and Zhou, Jinxin and Li, Xiao and You, Chong and Sulam, Jeremias and Qu, Qing},
title     = {A Geometric Analysis of Neural Collapse with Unconstrained Features},
booktitle = {Advances in Neural Information Processing Systems (NeurIPS)},
volume    = {34},
year      = {2021},
publisher = {Curran Associates, Inc.}
}

\end{document}